\documentclass[english,10pt]{article}

\usepackage{graphicx}
\usepackage{multirow}
\usepackage{mathptmx}
\usepackage{helvet}
\usepackage{courier}

\usepackage[letterpaper]{geometry}
\usepackage{setspace}
\usepackage{babel}
\usepackage{natbib}
\usepackage[usenames,dvipsnames]{xcolor}
\usepackage{subcaption}
\usepackage{tikz}
\usepackage{url}

\newcounter{words}

\newcommand*{\code}{\fontfamily{pcr}\selectfont}
\def\checkmark{\tikz\fill[scale=0.4](0,.35) -- (.25,0) -- (1,.7) -- (.25,.15) -- cycle;}

\graphicspath{{./figures/}}

\begin{document}

\newcommand{\reportcount}[1]{}

\newcommand{\textdegree}{$^\circ$}

\title{Automating Image Analysis by Annotating Landmarks with Deep Neural Networks}

\author{Mikhail Breslav$^{1*}$, Tyson L. Hedrick$^{2*}$, Stan Sclaroff$^{1}$, Margrit Betke$^{1}$} 
\maketitle

Running head: Automatic Annotation of Landmarks

$^1$ Department of Computer Science, Boston University, Boston, MA 02215 \\ 
\indent $^2$ Department of Biology, University of North Carolina at Chapel Hill, Chapel Hill, NC 27599

$*$ Author for correspondence (breslav@bu.edu,thedrick@bio.unc.edu)

Keywords: automatic landmark localization, annotation, pose estimation, deep neural networks, hawkmoths

\tableofcontents

\newpage

\onehalfspacing


\section{Abstract}

Image and video analysis is often a crucial step in the study of animal behavior and kinematics. Often these analyses require that the position of one or more animal landmarks are annotated (marked) in numerous images. The process of annotating landmarks can require a significant amount of time and tedious labor, which motivates the need for algorithms that can automatically annotate landmarks. In the community of scientists that use image and video analysis to study the 3D flight of animals, there has been a trend of developing more automated approaches for annotating landmarks, yet they fall short of being generally applicable. Inspired by the success of Deep Neural Networks (DNNs) on many problems in the field of computer vision, we investigate how suitable DNNs are for accurate and automatic annotation of landmarks in video datasets representative of those collected by scientists studying animals.

	Our work shows, through extensive experimentation on videos of hawkmoths, that DNNs are suitable for automatic and accurate landmark localization. In particular, we show that one of our proposed DNNs is more accurate than the current best algorithm for automatic localization of landmarks on hawkmoth videos. Moreover, we demonstrate how these annotations can be used to quantitatively analyze the 3D flight of a hawkmoth. To facilitate the use of DNNs by scientists from many different fields, we provide a self contained explanation of what DNNs are, how they work, and how to apply them to other datasets using the freely available library Caffe and supplemental code that we provide.


\reportcount{Abstract}


\section{Introduction}

\begin{figure*}[!ht]
\centering
	\includegraphics[width=1.00\linewidth]{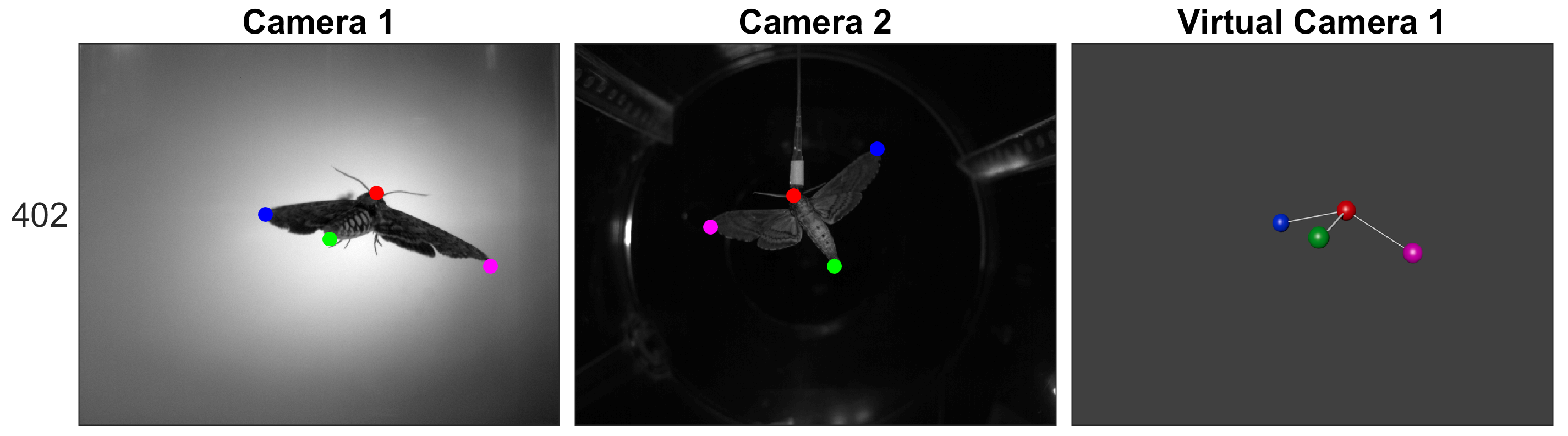} \\
	\caption[Example of obtaining 3D quantities form video data.]{\textbf{Example of obtaining 3D quantities from video data}. Four landmarks from a hawkmoth are annotated in two images taken simultaneously by two calibrated cameras, denoted Camera 1 and Camera 2. The four landmarks consist of the head (red circle), abdomen tip (green circle), left wing tip (blue circle), and right wing tip (magenta circle). After processing, these annotations are used to estimate the 3D positions of the four landmarks, shown in the right-most plot.}
\label{fig:intro}
\end{figure*}

Image and video analysis is often an essential component in the study of animal behavior and kinematics. One important use case of video analysis is in the estimation of 3D quantities of an animal, which typically requires that one or more landmarks from the body of the animal are annotated (marked) in video recordings from multiple cameras. Figure \ref{fig:intro} illustrates this process where landmarks from a hawkmoth are annotated and then used to estimate the 3D configuration of a hawkmoth. From left to right, the first two plots in Figure \ref{fig:intro} show landmark annotations for four different landmarks, and the third plot shows the estimated 3D positions of the landmarks. 

In this work, we focus on the general problem of \textit{automatic} annotation of landmarks, which we apply to the study of the 3D flight of hawkmoths. Ideally, an automatic landmark annotation method should take an image as input and subsequently output the 2D image location(s) of one or more landmarks. In the analysis of animal flight, annotating landmarks has traditionally involved manual labor from one or more people. In the case of annotating numerous landmarks in hundreds or thousands of frames across multiple cameras, this process can become laborious and time intensive, which motivates the need for automatic methods. Existing approaches to landmark annotation range in how automated they are. In the work of \cite{Tobalske2007}, landmarks on a hummingbird were marked with white paint and subsequently manually annotated in video data. \cite{Shelton2014} manually annotated natural landmarks on cliff swallows. \cite{Bergou2011} reduced the amount of annotation necessary by manually annotating landmarks of bats in an initial set of frames and then using tracking algorithms, which can be corrected with user input, to follow and annotate the landmarks over time. \cite{Ortega-Jimenez2014} achieved a mostly automated approach to landmark annotation by setting the view point of the camera in a way that facilitates the use of simple image processing steps for locating parts of a moth.





Some alternative approaches obtain 3D flight kinematics without the need to annotate landmarks. These approaches rely on a 3D graphics model of the animal and use a ``registration'' method to align the model to 2D image data. \cite{Fontaine2009} built a 3D graphics model of a \textit{Drosophila} fly and estimated its 3D motion by aligning the model to 2D image features. In the work by \cite{Breslav2014}, a 3D graphics model of a \textit{Tadarida brasiliensis} bat was built and used along with a Markov Random Field to find a 3D flight sequence that most agreed with the image data. The reliance of these approaches on accurate 3D graphics models hinders their use in other domains where such models are not readily available or are too costly to construct.

In the field of computer vision, researchers have extensively studied problems that require automatically annotating landmarks (also referred to as localizing landmarks) in images and videos. Many works in computer vision focus on solving these problems in videos of people. \cite{Zhu2012}, for instance, proposed an approach for localizing facial landmarks. \cite{Felzenszwalb2005} proposed an approach to the problem of 2D pose estimation of humans where the goal is to estimate the 2D position, rotation, and scale of various body parts. More recently, deep neural networks (DNNs) have gained fame in computer vision for producing top results on a variety of tasks including: object recognition and image classification \citep{Krizhevsky2012,Simonyan2014}, landmark localization \citep{Zhang2014}, and 2D pose estimation \citep{Toshev2014}. 

Inspired by the success of DNNs, our work aims to evaluate how suitable DNNs are for accurate and automatic localization of landmarks in the challenging case where the amount of labeled training data is small, as is typical in video data collected for analysis of animals. This is in contrast to the typical usage of DNNs in computer vision where large labeled training sets exist; see, for example, ImageNet \citep{Deng2009}, which contains more than 1 million images, each annotated with bounding boxes for all objects in the image. 
To facilitate our study we perform experiments on published high speed hawkmoth (\textit{Manduca sexta}) video data \citep{Breslav2016,Ortega-Jimenez2014}. \\

\noindent The remaining sections in this paper address the following questions:
\begin{itemize}
\item What are DNNs and how do they work? (Materials \& Methods)
\item What kind of DNN design can be used to automatically annotate landmarks? (Materials \& Methods) 
\item What decisions and considerations must take place before training a DNN?  (Materials \& Methods)
\item What level of performance can be obtained using DNNs for automatic landmark localization in hawkmoth videos? (Results)
\item How is the level of performance impacted by various factors including: dataset augmentation, network architecture, parameter values, and more? (Results)
\item How do I go about using DNNs for my own data/application? (Appendix B). 
\end{itemize}

\section{Materials and Methods}

\subsection{Background}

In this subsection we provide a brief introduction to several aspects of \textit{deep} neural networks (DNNs), which are foundational to our work. For a more comprehensive introduction to DNNs, see for example: \cite{Goodfellow2016}.

\begin{figure*}[htb]
\begin{subfigure}[t]{0.65\textwidth}
	\includegraphics[width=0.9\linewidth]{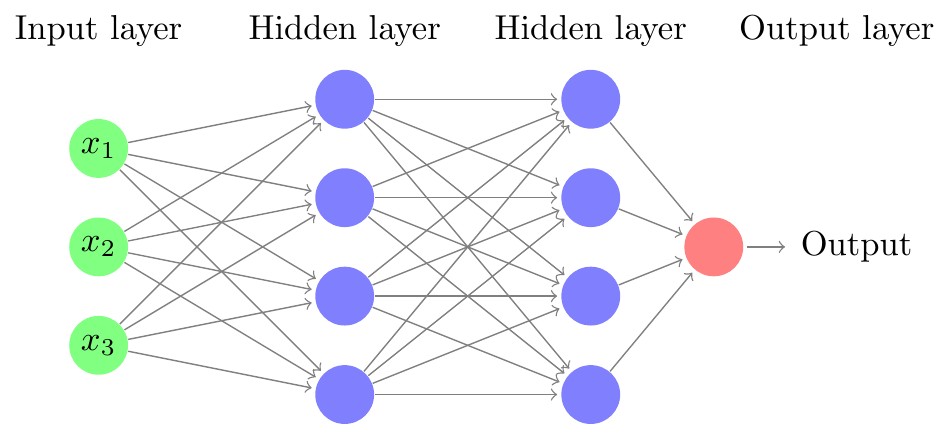}
	\caption{}
\end{subfigure}
\begin{subfigure}[t]{0.30\textwidth}
	\includegraphics[width=1.0\linewidth]{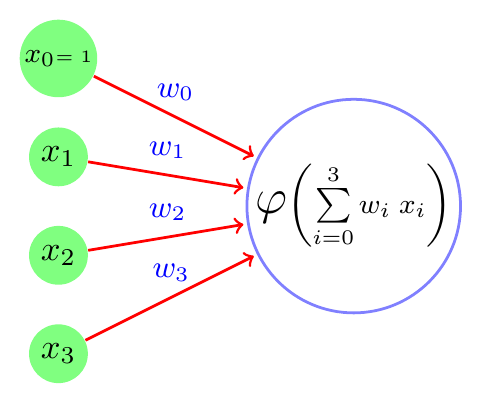}
	\caption{}
\end{subfigure}
\caption[Example neural network.]{\label{fig:nn} \textbf{An example neural network.}  \textbf{(a)} This example neural network of depth 3 contains two hidden layers and has an output layer with a single output neuron. Most generally a neural network can have many hidden layers, each containing any amount of neurons. The output layer can also have any number of neurons (outputs). Gray edges represent the weighted connections in the network. The bias terms are not shown for visual clarity but would normally be connected to all neurons. \textbf{(b)} This is an illustration of what an individual artificial neuron does. The artificial neuron denoted by the blue circle computes a linear combination of 3 inputs ($x_1,x_2,x_3$), weighted by the weights ($w_1,w_2,w_3$) and adds a bias term $w_0 \times x_0 = w_0$. The resulting sum is then input to a non linear function $\varphi$ that computes a final value which is the output of the artificial neuron.}

\end{figure*}

\subsubsection{Deep Neural Networks} 
To explain what a \textit{deep} neural network is, we begin by introducing neural networks. A neural network is a model that can be trained to predict a label for a given input. Commonly, neural networks are used to perform tasks such as classification where the predicted label is discrete valued, and regression where the predicted label is continuous valued. A neural network is typically composed of an input layer, one or more hidden layers, and an output layer. Figure \ref{fig:nn}a shows an example neural network which has two hidden layers. The input layer represents the input to the network which in the context of images is often the raw pixel values of an image. The hidden layers consist of artificial neurons which collectively map the input to an intermediate representation known as a ``feature'' vector. The output layer also consists of artificial neurons and it maps these feature vectors to one or more outputs. The depth of a neural network can be defined as the length of the shortest path from the input to the output, which is equal to the number of hidden layers plus one for the output \citep{Bengio2009}. A \textit{deep} neural network is then defined as a neural network whose depth is large.\footnote{The minimum depth for a network to be considered \textit{deep} is not well defined in the literature \citep{Schmidhuber2015}.} Each artificial neuron in the network contributes to the overall mapping by computing a weighted combination of its inputs, summed together with a bias term, and evaluated with a non linear function. Figure \ref{fig:nn}b illustrates the action of a single artificial neuron, denoted with a blue circle. The neuron (we drop the qualifier artificial for brevity) computes a weighted sum over 3 inputs $(x_1,x_2,x_3)$, and adds a bias term represented as $w_0 \; x_0 $, with $x_0 = 1$. The final output of the neuron is $\varphi \left( \sum\limits_{i=0}^3{w_i \; x_i}\right)$, where $\varphi$ is a non linear function. More generally a neuron can take $n$ inputs which will either be the input to the network or the output of a hidden layer. Note that in Figure \ref{fig:nn}a the gray edges represent the weights of the network. The biases, which are not shown for visual clarity, should be connected to each neuron.

Given a neural network with initial values for the weights and biases, an input sample can be ``forward propagated'' through the network. In forward propagation, each neuron performs a computation on its inputs before neurons in subsequent layers perform theirs. The process ends once the output is computed. To produce meaningful predictions, a neural network must be trained. The goal of training is to modify the initially set weights and biases of a neural network so that it predicts labels as accurately as possible for a set of labeled input samples known as the training set. To evaluate how well a network performs prediction on a training set, a ``loss function'' is defined. The loss function, or ``loss'' for short, is a measure of how much the predictions for a training set differ from the true labels and it is a function of the weights and biases of the network. Generally the loss function is designed to reflect the loss averaged over the training set. In our work we will refer to the evaluation of a loss function on the training set, as the ``training loss'', and the evaluation of a loss function on the testing set, as the ``test loss'' or ``testing loss''.

From an optimization point of view, the goal of training is to find the weights and biases that minimize the loss. The standard approach to this optimization is based on the method of gradient descent using back propagation \citep{Rumelhart1988}. In the gradient descent method, for a given function $F(\theta)$, where $\theta$ is a parameter vector and $F$ is differentiable and defined in a neighborhood of some parameter setting $\theta_{n}$, one can obtain a new function value $F(\theta_{n+1})$ where $F(\theta_{n+1}) \le F(\theta_{n})$. The new function value is obtained by updating the parameter vector so that it moves in the direction of the negative gradient of $F$ evaluated at the current parameter value. The update in the parameter vector is expressed as  $\theta_{n+1} = \theta_{n} - \eta \nabla F(\theta_{n})$, where $\eta$ is a small scalar. By repeatedly updating the parameter vector in this way the gradient descent method will converge to a local minimum of $F$, which is also the global minimum if $F$ is convex. In the context of neural networks, the function $F$ is the loss function, and the parameter vector $\theta$ is the weights and biases of the network. To compute the gradient of the loss function with respect to the weights and biases of the network, back propagation is used which first computes gradients with respect to weights and biases at the end of the network and then uses the ``chain rule'' from calculus to compute gradients with respect to the weights and biases in earlier layers. Traditional gradient descent requires that the whole training set is forward propagated before the loss function is evaluated and a single update to the weights and biases is performed, which in practice is expensive since datasets can be large. Instead, stochastic gradient descent (SGD) or mini-batch SGD are used, where the loss function is evaluated on either a single training sample or a subset of training samples, and an update to the weights and biases can be performed much quicker. A more formal discussion on gradient descent and SGD is provided in the work of \cite{Lecun2012}.

\begin{figure*}[htb]
\includegraphics[width=1.0\linewidth]{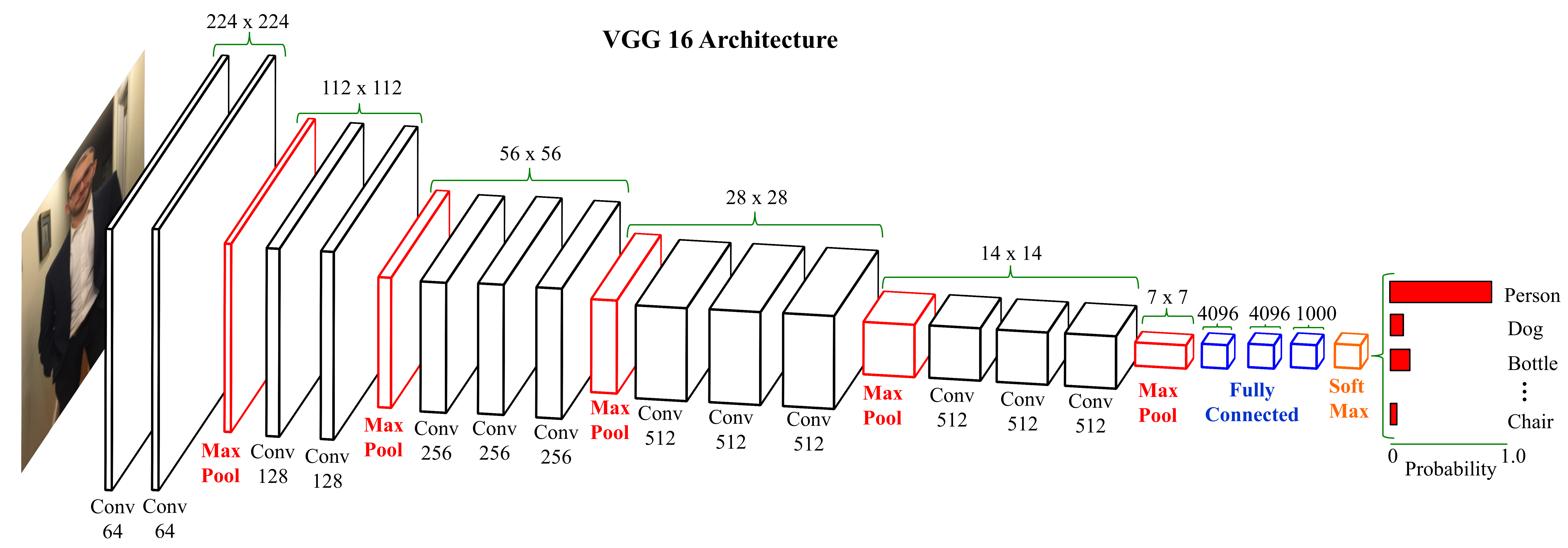}
\caption[Illustration of the VGG 16 network architecture.]{\textbf{Illustration of the VGG 16 \citep{Simonyan2014} network architecture, layer by layer}. The VGG 16 network consists of 16 layers that have weights and biases. These layers include convolutional layers (``Conv'') denoted by black rectangular volumes and fully connected layers denoted by blue rectangular volumes. The height and width of each convolutional layer is specified above the layer and the depth is specified below the layer. Max pooling layers, denoted as red rectangular volumes, have the same depth as the layer that precedes them, and their height and width is the same as the layer that succeeds them. The fully connected layers are 1D columns of neurons with the number of neurons specified above the layer. A soft max layer denoted as an orange rectangular volume is used to convert the output of the neurons in the last fully connected layer to class probabilities. The weights and biases of the network are not explicitly shown. A more in depth explanation of the VGG 16 network and its layers is given in Appendix A. The VGG 16 network was originally trained for the task of image classification. Here we illustrate an example classification where the network is given an input image of a person and the output probability distribution is largest for the output corresponding to the label ``person''.}
\label{fig:vgg16}  
\end{figure*}

\subsubsection{Deep Neural Network Architecture}
The neural network in Figure \ref{fig:nn}a has a ``shallow'' (not deep) architecture, which is useful as an illustration, but is not representative of the kinds of networks being used in practice. Instead, the networks used to solve many real world problems have deeper and more complex architectures. Theoretical results support the idea that deeper architectures can more efficiently represent a function as compared to shallower ones \citep{Bengio2009}. Today, researchers in computer vision and machine learning continue to design and study DNN architectures with the goal of obtaining the top results on a variety of problems.

One type of DNN, known as a Convolutional Neural Network (CNN or ConvNet), has succeeded on various computer vision problems by leveraging a particular architecture. The architecture of a CNN uses ``convolutional'' layers, which perform the mathematical operation of convolution. The use of convolution is motivated by the idea that local patterns in an image are informative and that informative local patterns should influence the network regardless of their absolute location in an image \citep{Lecun1998}. Convolutional layers are also beneficial in that they reduce the total number of weights and biases in the network when used instead of traditional layers, which are also called fully connected layers. One of the earliest successes of CNNs was in handwritten character recognition \citep{Lecun1998}. More recently \cite{Simonyan2014} proposed 16 and 19 layer CNNs, called VGG 16 and VGG 19, which achieved the best results in 2014 on the task of classifying and locating objects in images. Figure \ref{fig:vgg16} shows the layer by layer composition of the VGG 16 network, and illustrates what a reasonable network output would be for an input image of a person. Unlike the neural network introduced in Figure \ref{fig:nn}a, where all layers containing neurons are arranged as 1D columns, the VGG 16 network has layers consisting of 3D volumes (width x height x depth) of neurons. These layers are the convolutional layers and the max pooling layers, which are labeled in Figure \ref{fig:vgg16} with `Conv' and `Max Pool', respectively. As before neurons in the same layer do not connect to each other. The VGG 16 network plays an important role in our work because we use architectures that are directly based on this network. For a more in depth explanation of the different types of layers used in the VGG 16 network, which are also common to many other CNNs, see Appendix A. 


\subsection{Experimental Setup}

In this section we describe some of the key parameters that need to be considered and set prior to training and testing a DNN. The goal of our experiments is to obtain quantitative results detailing how different values and settings of these parameters impact landmark localization accuracy. Preliminary experiments were performed to find an initial set of parameter values that resulted in a reasonably low test loss, indicating that the network learned sufficiently well to generalize. Many parameter values could be quickly rejected when the training loss was observed to either diverge or to decrease too slowly from a large value. We will refer to the parameter values found in our preliminary experiments as the ``default'' parameter values. These default parameter values are also used as a guide for what parameter value ranges we consider in our experiments. Exploring how all parameters jointly influence landmark localization is a combinatorial problem and is not feasible to perform due to the large amount of time (tens of hours) it takes to train a \textit{single} DNN. Instead, our experiments evaluate how deviations from default parameter values influence results\footnote{Alternative approaches for exploring parameter values are discussed in \cite{Bengio2012}.}. In the rest of this subsection, we will describe what the parameters are and, where appropriate, we will state what the default value is. Conceptually our experiments can be thought of as an exploration of the parameter space around a point (default parameter values), exploring one dimension at a time. 

\subsubsection{Datasets}
For our experiments we use hawkmoth video data from \cite{Ortega-Jimenez2014}. The video data captures an individual hawkmoth (\textit{Manduca sexta}), from multiple cameras, while it hovers in a vortex chamber where the wind speed is high. Specifically, we use videos obtained from cameras 1 and 2, which simultaneously recorded the flight of the hawkmoth. Cameras 1 and 2, which are identical, are equipped with a 28 mm lens, record at 400 frames per second and have a resolution of 600 x 800 pixels. Both of these cameras have been calibrated \citep{Theriault2014}, so their relative positions and orientations in 3D are known. Landmark annotations for 800 frames in each video were obtained by using annotations published by \cite{Breslav2016} and performing additional annotations ourselves. Landmark annotations consist of the 2D image position of the head, abdomen tip, left wing tip, and right wing tip. We make the hawkmoth data used in our experiments freely available along with landmark annotations\footnote{\url{http://www.cs.bu.edu/~betke/research/HRMF2/}}.\\

\subsubsection{DNN Architecture}
The DNN architectures used in our experiments are directly based on the VGG 16 network of \cite{Simonyan2014}. The default network architecture used in our experiments is a network we call ``VGG 7 + FC8''. VGG 7 is what we call the network obtained by taking VGG 16 and removing everything after the max pooling layer that follows the $7^{th}$ convolutional layer; counting starts from the layer closest to the input. On top of VGG 7 we added a single fully connected layer consisting of 8 neurons (FC8), which produces the output of the network. The quantity of 8 neurons is chosen so that the network outputs a pair of $(x,y)$ image locations for four hawkmoth landmarks of interest: the head, abdomen tip, left wing tip, and right wing tip. 

In our experiments we also investigated the performance of architectures similar to VGG 7 + FC8, but with different depths.  We will refer to these alternative architectures as VGG $X$ + FC8, where VGG $X$ corresponds to the subset of VGG 16 that remains when removing everything after the max pooling layer that follows the $X^{th}$ convolutional layer.

%
\subsubsection{Network Initialization}

There are two primary ways to initialize the weights and biases of a network. The first way initializes the weights and biases to values that were already learned from training the same network, or a superset\footnote{Here we define the superset of a network $a$ as another network $b$ that contains $a$ as part of its architecture.} of it, on a different dataset. A network initialized using this approach is said to be ``pretrained''. The second way is to initialize the weights and biases manually by choosing a constant or generating values randomly from a distribution. Training a network that has been initialized using the second approach is also referred to as training from ``scratch''. 

In our experiments the first approach is the default way we initialized an architecture. Specifically, for an architecture VGG $X$ + FC8, we initialized the weights and biases of the VGG $X$ portion of the network to the values learned from training VGG 16 on ImageNet, which are freely available for download\footnote{http://www.robots.ox.ac.uk/~vgg/research/very\_deep/}. For the weights and biases of the fully connected portion of the network (FC8), which is not part of VGG 16, a constant of 0 was used. 

We also performed experiments to evaluate how our network performed when trained from scratch. In these experiments the weights were initialized either from a Gaussian distribution (0 mean, 0.01 standard deviation), or using ``xavier'' initialization, where the weights are drawn from a distribution whose variance is determined by the number of inputs and outputs a particular neuron has \citep{Glorot2010}. 

\subsubsection{Finetuning}
Our default network architecture VGG 7 + FC8 was chosen so that the VGG 7 portion of the network can be pretrained and used as a feature extractor, while the FC8 part of the network performs linear regression and needs to be trained from scratch. As a result, for most experiments training only involves the fully connected layer. Finetuning, however, is an approach where pretrained layers can be further trained to ``tune'' the network for a particular dataset. In our case, when we perform finetuning all layers of the network are trained. It is worth noting that in libraries like Caffe \citep{Jia2014} a learning rate can be specified for each layer of the network. In our experiments we investigated the impact of finetuning VGG 7 + FC8. 

\subsubsection{Dataset Augmentation}
Data augmentation is an approach used for increasing the size of a training set by applying transformations to images in the training set \citep{Simard2003,Krizhevsky2012}. In our experiments we investigated how data augmentation using combinations of translation, rotation, and scale, impact network performance. Additionally, we also investigated how the amount of data augmentation performed, which determines the total number of training samples available, influences performance. Our default data augmentation uses translation alone and results in 200,000 training samples in total. As a frame of reference we also evaluated network performance when no data augmentation is performed.

Translational data augmentation was implemented by taking a 400 x 600 pixel crop from an original 600 x 800 pixel image and subsequently resizing the crop to 224 x 224 pixels. The 400 x 600 pixel crops are taken so that the position of the hawkmoth inside the crop was distributed uniformly at random. The crop was constrained to contain the full hawkmoth body. A bounding box of the hawkmoth body was generated from the hawkmoth segmentations provided by \cite{Breslav2016}. Rotational data augmentation was implemented by rotating the hawkmoth around the center of its bounding box, randomly in the range of -45 degrees to 45 degrees. Scaling data augmentation was implemented using the ``imresize'' function in Matlab which performs downsampling or upsampling to the image using bilinear interpolation. The amount of scaling was chosen randomly in the range of 0.5 to 1.5. 

\subsubsection{Deep Learning Library}
To train and evaluate DNNs we used the publicly available deep learning framework Caffe, developed by \cite{Jia2014}. The Caffe library implements mini-batch SGD \citep{Lecun2012} using back propagation and is able to leverage GPU resources. See Appendix B for an overview of how to use Caffe to train and test DNNs. We provide free software to help facilitate training and testing of DNNs with Caffe\footnote{\url{http://www.cs.bu.edu/~betke/research/ALADNN/}}. 
 
\subsubsection{Learning Rate} The learning rate is one of the parameters that greatly impact training of a DNN. The value of the learning rate determines how large of a step is taken during SGD. In the expression $\theta_{n+1} = \theta_{n} - \eta \nabla F(\theta_{n})$, $\eta$ is the learning rate. In our experiments we use a default learning rate of $100 \times 10^{-12} = 10^{-10}$ for training the fully connected layer (FC8). Our experiments investigating the influence of network architecture and network initialization on DNN performance also consider a range of learning rates. 

\subsubsection{Batch Size} The batch size is the number of training samples used for a single iteration of training which results in a single update of the weights and biases in the network. The loss function is evaluated on the batch of training samples and as a result subsequent gradient computations depend on these samples. The default batch size we used is 32. Additional experiments were performed to evaluate the influence of batch size on network performance.

\subsubsection{Training Iterations} The number of training iterations determines in combination with batch size how many training sample 
are seen during training, which also determines the amount of time training takes. The default number of training iterations we performed is 10,000. Additional experiments were performed to evaluate how the number of training iterations impacts performance.

\subsubsection{Training and Testing Split} For our experiments (not including multi-view experiments) we had available a total of 800 consecutive frames which were annotated. By default we used the first 400 frames for training (not including any data augmentation performed) and the latter 400 frames for testing. We also performed two additional experiments to study alternative training and test splits. The first experiment studies the impact of making the training and testing sets more similar to each other. This is accomplished by using odd numbered frames for training and even numbered frames for testing. We refer to this alternative split as ``interleaved''. The second experiment studies how different sizes of the training set (prior to data augmentation) impacts performance on a test set. Specifically, we used 200, 400, and 600 training images and a random test set of 200 frames. 

\subsubsection{Evaluation} After a DNN is trained its weights and biases are finalized, and the network can be used on new inputs. For evaluation a set of test images (inputs) are forward propagated through the network and a loss function is evaluated. In all of our experiments we used the loss function defined by: $\frac{1}{n} \; \sum\limits_{i=1}^n ||\mathbf{y_i}-\mathbf{y_i^*}||^{2}$, where: $i$ denotes the $i^{th}$ test sample out of $n$, $\mathbf{y_i}$ is the output of the network on the $i^{th}$ test sample represented as an 8 dimensional vector, and $\mathbf{y_i^*}$ is the 8 dimensional vector consisting of ground truth annotations for the $i^{th}$ test sample. We will refer to this loss as the \textit{squared loss} for convenience, though more accurately the loss reflects the expectation (average) of the squared loss due to the $\frac{1}{n}$ term. In all of our experiments we evaluated the performance of a DNN by evaluating this loss function on the test set. Recall, that this is the same loss function that is integral in training and influences the derivatives computed during backpropagation. 

For some of our experiments we also report the more interpretable mean absolute error (MAE), which represents the average distance between a predicted landmark location and its corresponding ground truth location. The MAE can be defined as follows: let $y_i$ be the 2D image location of a particular landmark in the $i^{th}$ test image, let $y_i^*$ be the ground truth 2D image location of that landmark in the same test image, and let the error be defined as $e_i = ||\mathbf{y_i}-\mathbf{y_i^*}||$, then MAE is given by: $\frac{1}{n} \; \sum\limits_{i=1}^n |e_i|$.

It is also important that we clarify the resolution of the image upon which these evaluation metrics are computed. In all experiments where squared loss is reported, it is computed on images of size 224 $\times$ 224 pixels. The dimensions 224 $\times$ 224 represent the width and height of all images used for training DNNs in our experiments. Images of these dimensions are obtained from our original images of size 600 $\times$ 800 by taking a 400 $\times$ 600 crop, and then rescaling it to 224 $\times$ 224. The ground truth annotations (labels) associated with training images are also transformed to be consistent with these dimensions. As a result, after training, our DNNs will output landmark locations relative to an image with dimensions 224 $\times$ 224. All training and testing losses reported in our work were computed on images of this resolution. In contrast, in all experiments where MAE is reported, the MAE was computed on images at the original resolution of 600 $\times$ 800. 


\subsubsection{Multi-view} In our multi-view experiments we trained one DNN (DNN$_1$) using video from camera 1, and another DNN (DNN$_2$) using video from camera 2. Both DNNs were trained using our default split, where the first four hundred frames from a video are used for training and the last four hundred frames are used for testing. Once DNN$_1$ and DNN$_2$ are trained they can be used to predict landmark locations for their respective test sets. Given predicted landmark locations $\mathbf{y}_1^i$ for frame $i$ of video 1, and predicted landmark locations $\mathbf{y}_2^i$ for frame $i$ of video 2, 3D triangulation \citep{Hartley2004} can be used to estimate (``reconstruct'') the 3D position of the landmarks at time $i$, which can be thought of as a representation of the 3D pose of the hawkmoth at time $i$.

\section{Results}
\label{sec:results}

\noindent \textbf{\normalsize Default Network Parameters}\\
As we have noted earlier, our experimental results show how changes to parameter values influence network performance measured by the test loss, with a point of reference being a DNN trained using default parameter values. The default parameter values are compactly provided in Table \ref{tab:default_parameters}. Also recall that all test loss and training loss values reported in the results are computed on images with dimension 224 $\times$ 224, and all MAE values reported were computed on images with the original dimensions of 600 $\times$ 800. \\


\setlength{\tabcolsep}{5pt}
\begin{table}[th]
\centering
\caption[Default parameter values used for DNN experiments.]{\textbf{Default parameter values for:} the number of training samples (NTS), data augmentation type (DA), architecture (Arch.), base learning rate (BLR), learning rate multiplier for VGG layers (VGG LRM), learning rate multiplier for fully connected layer (FC LRM), pretraining of the VGG layers (VGG PT), number of training iterations (NI), batch size (BS), and finetuning (FT). \textbf{Note:} to be consistent with Caffe the learning rates for the VGG and FC layers are provided as constants that are relative to the base learning rate. The actual learning rate for the VGG and FC layers is obtained by taking the base learning rate and multiplying it by the VGG and FC multipliers respectively.}
\label{tab:default_parameters}
\def\arraystretch{1.5}
\begin{tabular}{|c|c|c|c|c|c|c|c|c|c|}

\hline
NTS & DA & Arch. & BLR & VGG LRM & FC LRM & VGG PT & NI & BS & FT \\
\hline
200,000 & T & VGG 7 + FC8 & $10^{-12}$ & 0 & 100 & \checkmark & 10,000 & 32 & X \\
\hline

\end{tabular}
\end{table}

\begin{figure*}[!ht]
\begin{subfigure}[!ht]{0.5\textwidth}
	\includegraphics[width=0.95\linewidth]{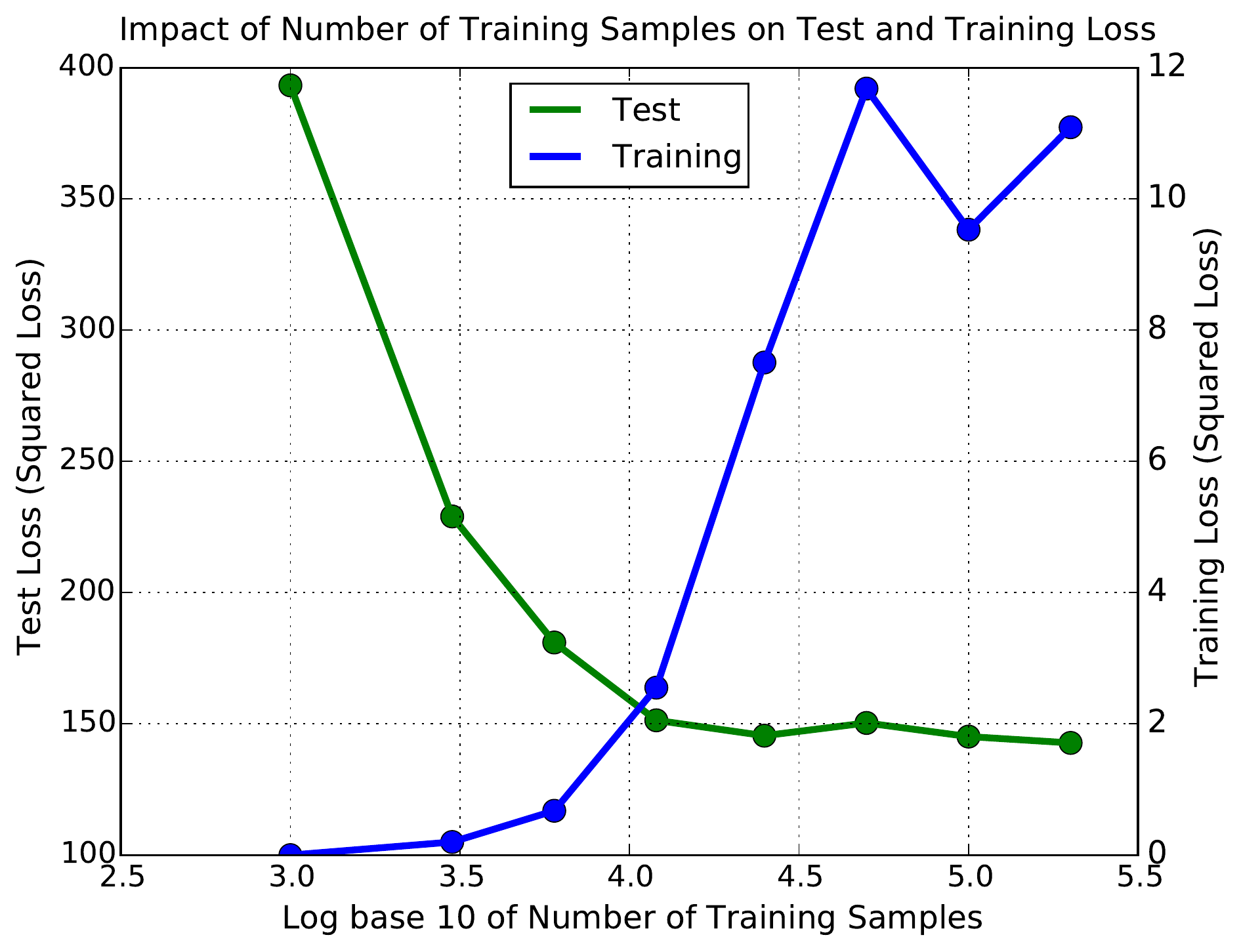}
	\caption{}
\end{subfigure}
\begin{subfigure}[!ht]{0.50\textwidth}
	\includegraphics[width=0.95\linewidth]{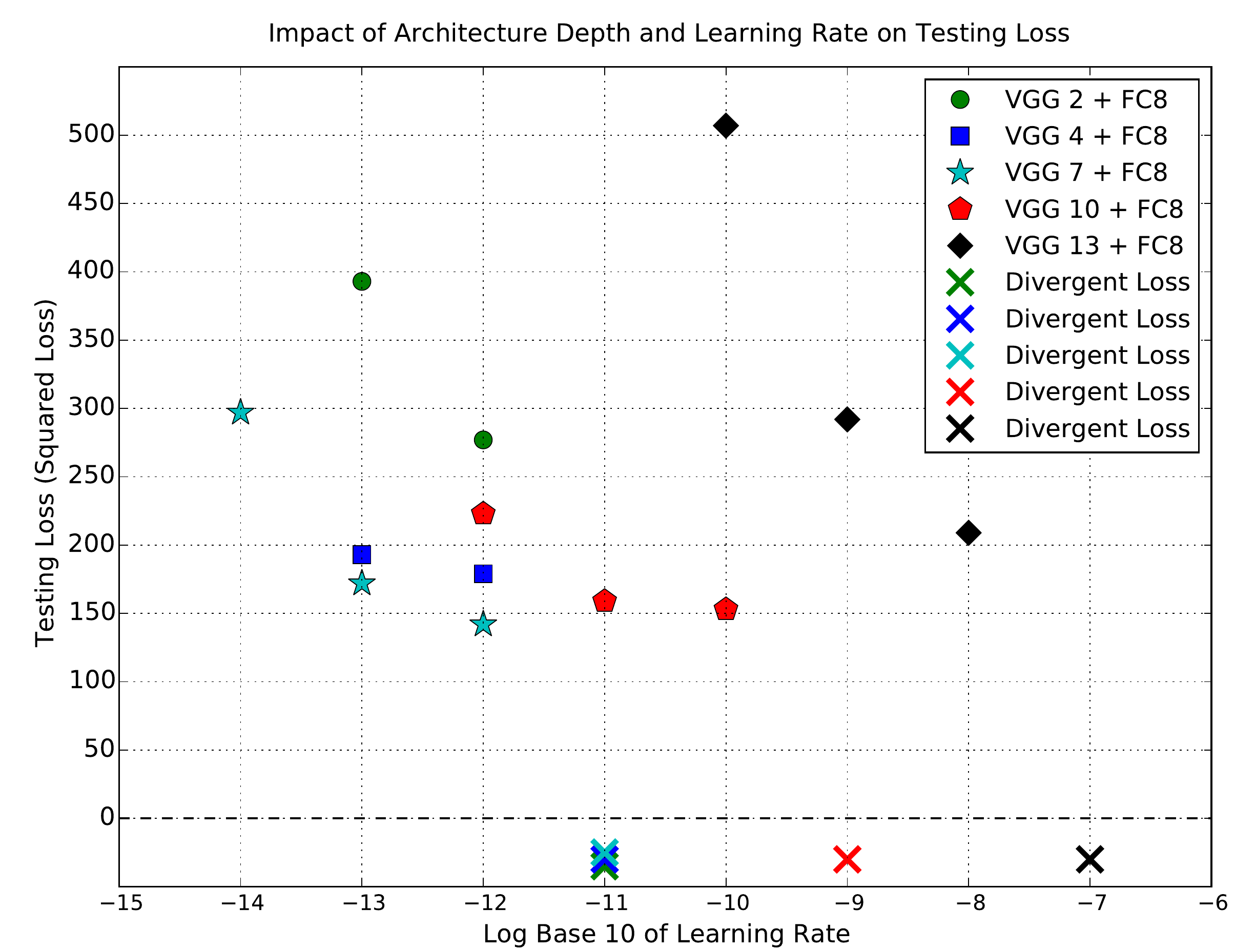}
	\caption{}
\end{subfigure}

\begin{subfigure}[!ht]{0.50\textwidth}
	\includegraphics[width=0.90\linewidth]{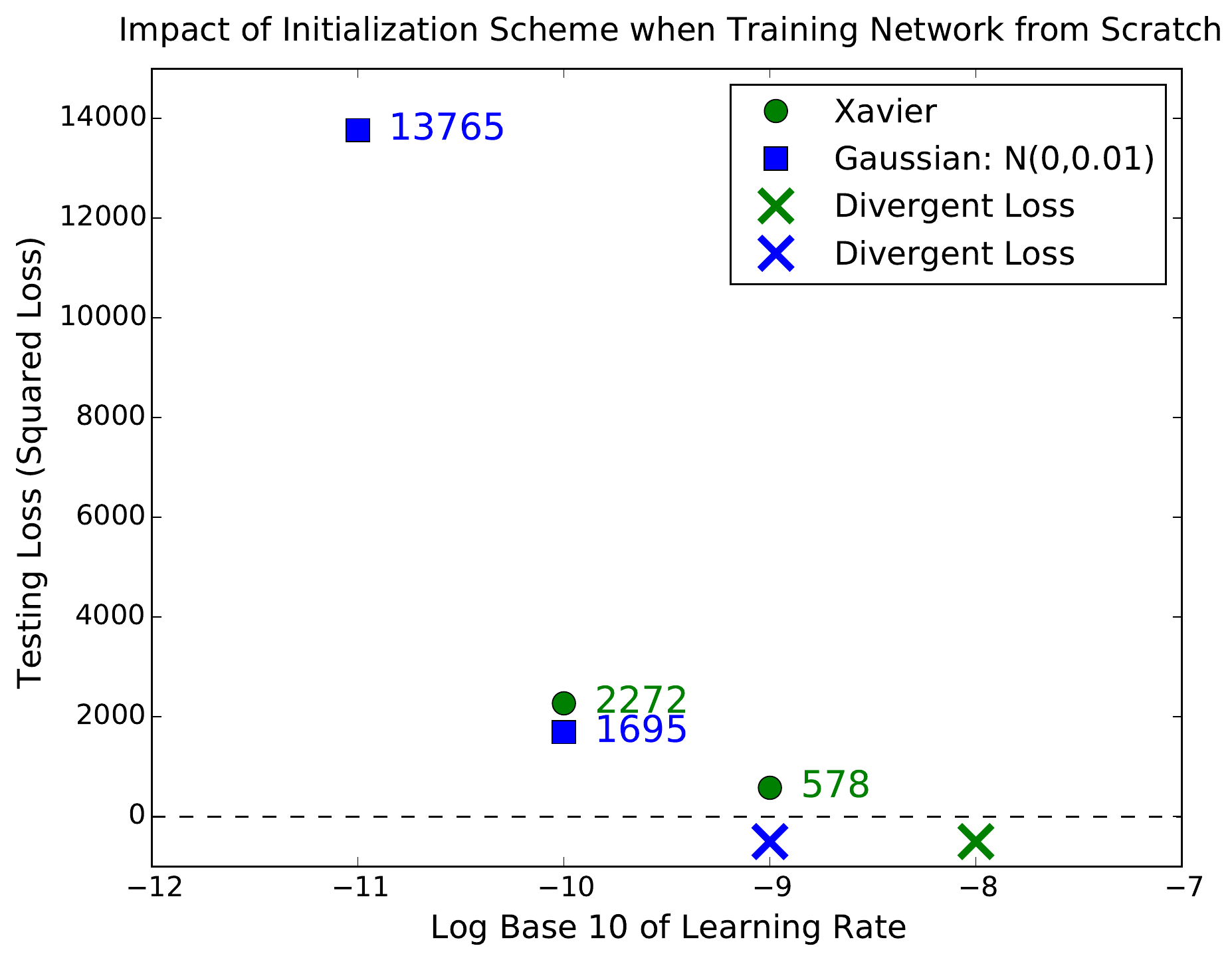}
	\caption{}
\end{subfigure}
\begin{subfigure}[!ht]{0.50\textwidth}
	\includegraphics[width=0.90\linewidth]{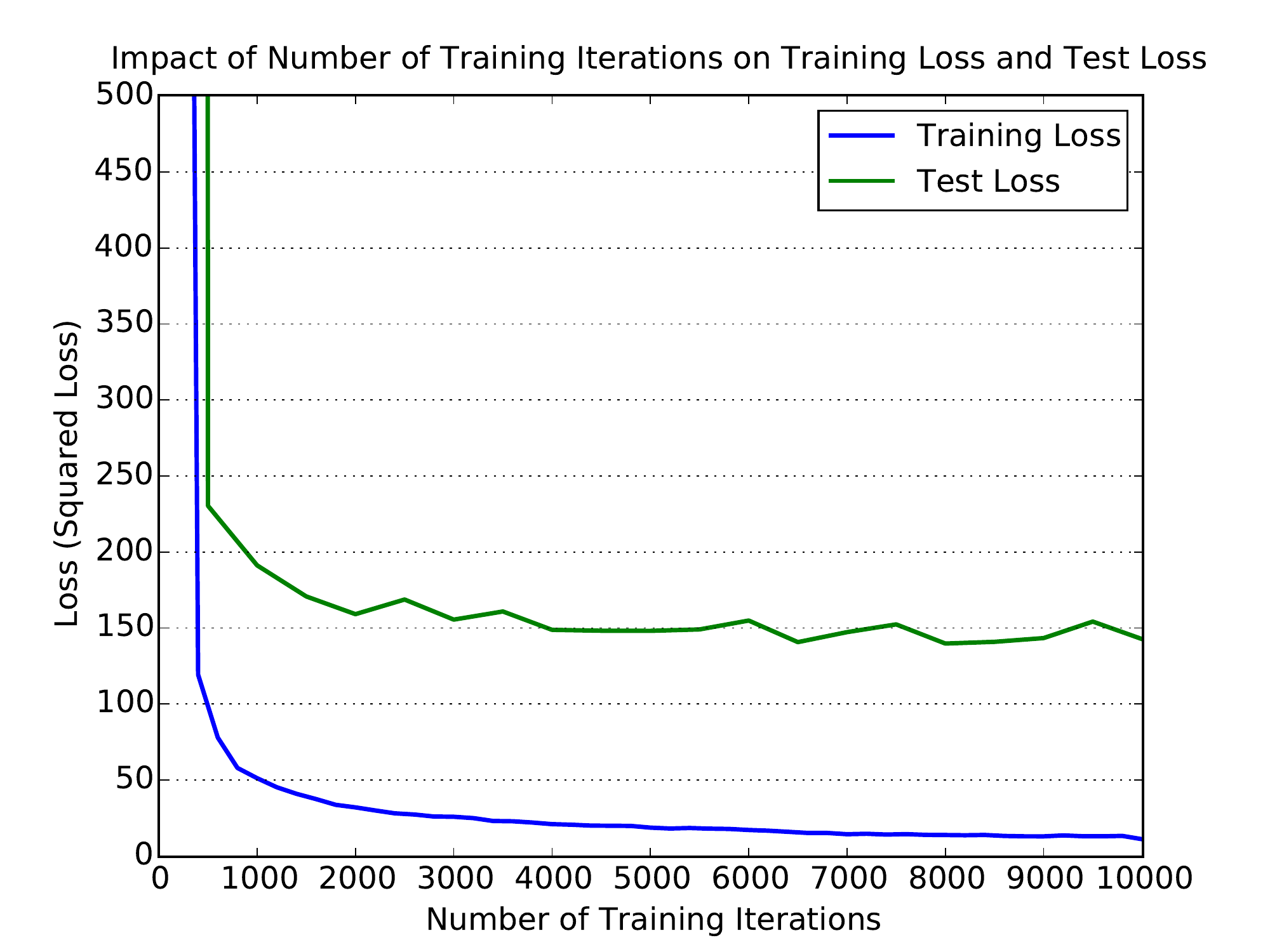}
	\caption{}
\end{subfigure}

\begin{subfigure}[!ht]{0.50\textwidth}
	\includegraphics[width=0.95\linewidth]{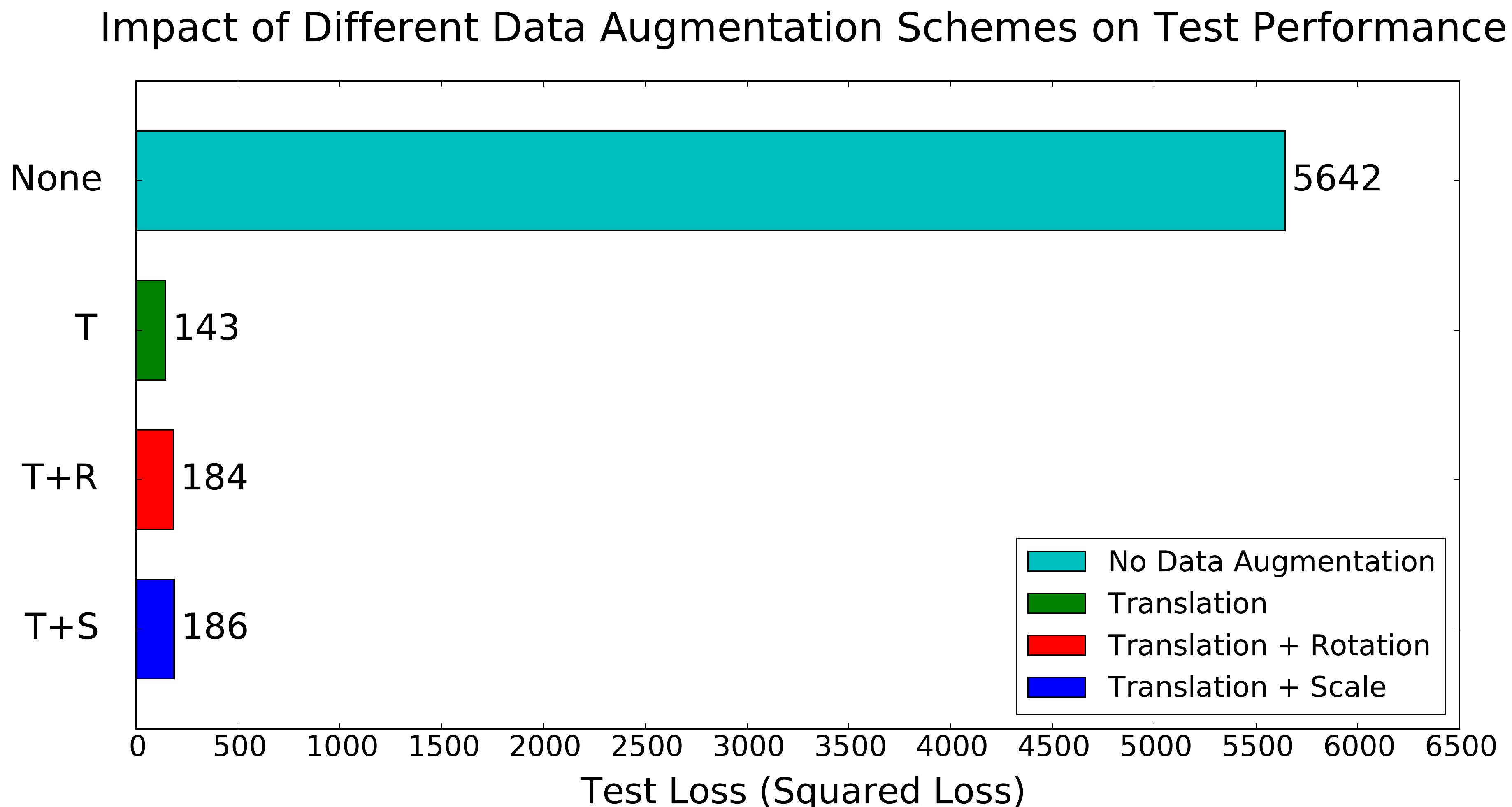}
	\caption{}
\end{subfigure}
\begin{subfigure}[!ht]{0.50\textwidth}
	\includegraphics[width=0.95\linewidth]{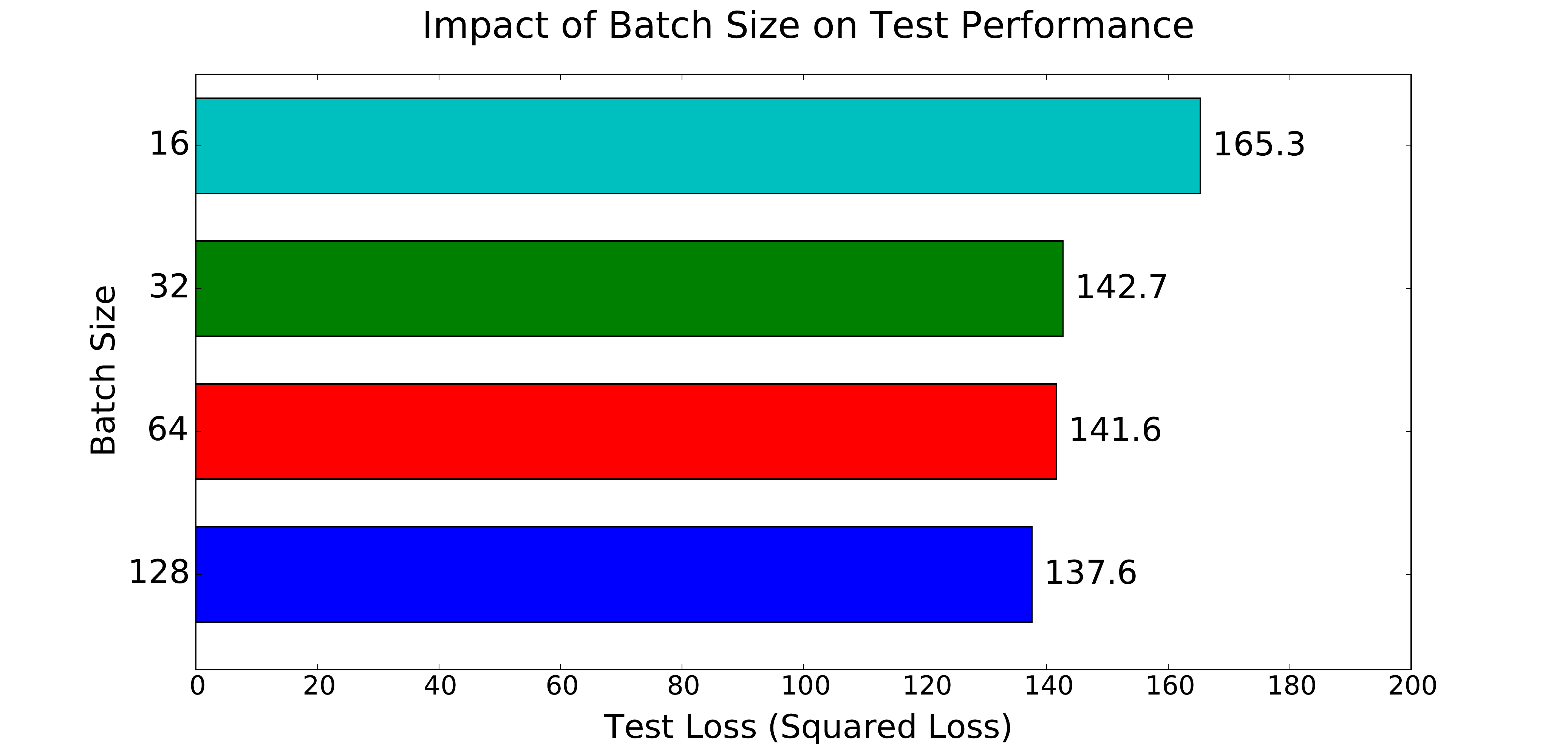}
	\caption{}
\end{subfigure}

\caption[Experimental results showing influence of different parameter values.]{\textbf{Experimental results show how different parameter values impact the test performance of the trained network.} These results summarize the quantitative impact of (a) the number of training samples, (b) different architectures and fully connected layer learning rates, (c) network initialization from scratch schemes, (d) number of training iterations, (e) data augmentation type, and (f) batch size.}
\label{fig:results}
\end{figure*}

\noindent \textbf{\normalsize Number of Training Samples:} Figure \ref{fig:results}a shows how the number of training samples used for training impacts both the final training loss (blue plot) and the final testing loss (green plot). Note the total samples seen during training is equal to the number of iterations performed multiplied by the batch size, which are both fixed. Thus, the total samples seen during training with default parameter values is $10,000 \times 32 = 320,000$. For the default training set size of 200,000 this means that each sample is seen at least once during training, but no more than twice. \\

\noindent \textbf{\normalsize Architecture Type:} Figure \ref{fig:results}b shows the impact of several different architectures on test performance. The architectures tested differ in the subset of VGG 16 used for feature extraction. The smallest network is VGG 2 + FC8 and the largest is VGG 13 + FC8. Each architecture is evaluated over several different learning rates for the fully connected layer. The crosses plotted below the $y=0$ line indicate the pairs of architectures and learning rates that resulted in the test loss \textit{diverging}. \\

\noindent \textbf{\normalsize Training from Scratch:} Figure \ref{fig:results}c shows how different initializations for all the weights and biases of the network impact the final testing loss. Each initialization scheme is evaluated for several fully connected layer learning rates. \\


\noindent \textbf{\normalsize Number of Iterations:} Figure \ref{fig:results}d shows how training loss (blue plot) and testing loss (green plot) vary with the number of training iterations performed. The training loss plotted reflects the average training loss over a window of 200 iterations, computed every 200 iterations. Testing loss was computed every 500 iterations.\\

\noindent \textbf{\normalsize Data Augmentation:} Figure \ref{fig:results}e shows how different data augmentation schemes influenced test performance. The green plot corresponds to data augmentation done using translations alone (denoted by T). The red plot corresponds to data augmentation done using both translation and rotation (denoted by T + R). The blue plot corresponds to data augmentation done using both translation and scale (denoted by T + S). In all cases data augmentation was used to increase the initial 400 training samples to 200,000 training samples. The final training set consists of 400 original training samples and 199,600 training samples generated from data augmentation. As a point of comparison, the black plot shows the performance when no data augmentation is performed. \\ 

\noindent \textbf{\normalsize Batch Size:} Figure \ref{fig:results}f shows how the batch size impacts test performance. An additional experiment was performed where the batch size was 8 which resulted in the loss diverging. \\

\noindent \textbf{\normalsize Finetuning:} Finetuning the default network resulted in a negligible difference in testing loss. One experiment performed finetuning by training all layers of the network with the VGG LRM set to 1 and resulted in a test loss of 142.75. A second finetuning experiment was performed where the VGG LRM was set to 10 and resulted in a test loss of 142.79. Increasing the BLR to $10^{-11}$ and setting both the VGG LRM and FC LRM to $1000$ resulted in a test loss of 514.66. The default network which does not use finetuning resulted in a loss of 142.7. \\

\begin{figure*}[!ht]
\centering
\begin{subfigure}[!ht]{0.80\textwidth}
	\includegraphics[width=1.0\linewidth]{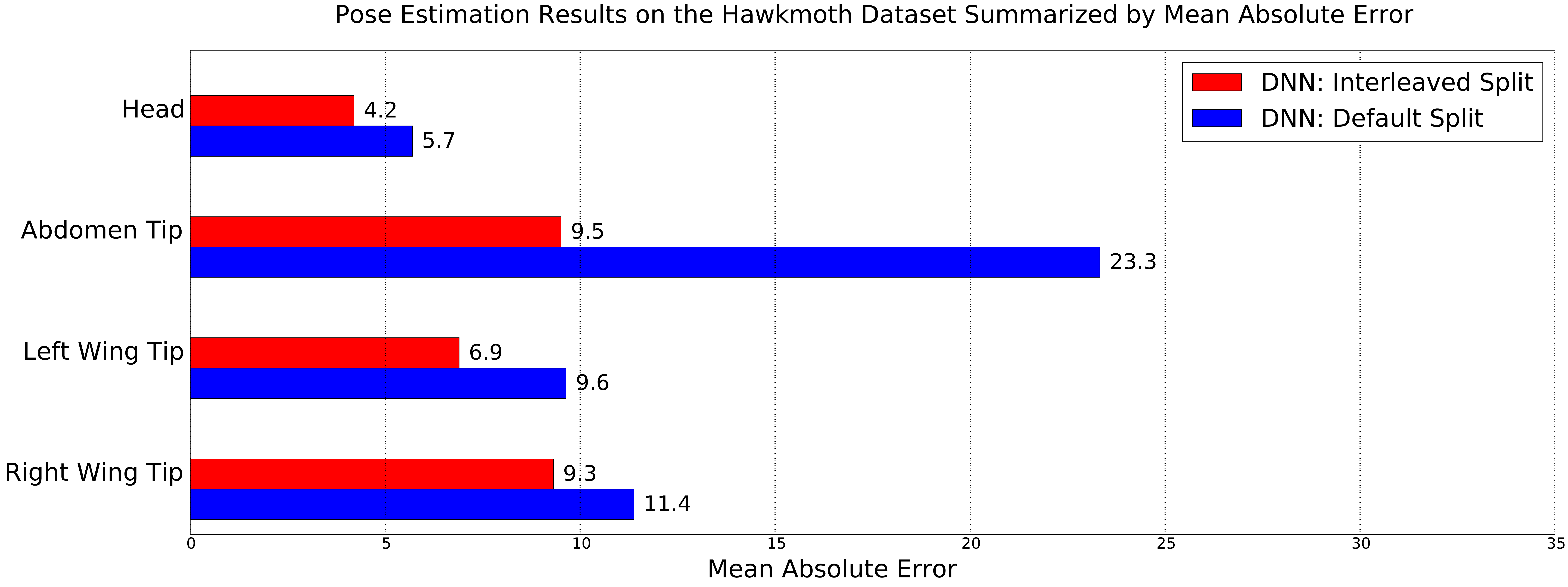}
	\caption{}
\end{subfigure}

\centering
\begin{subfigure}[!ht]{0.80\textwidth}
	\includegraphics[width=1.0\linewidth]{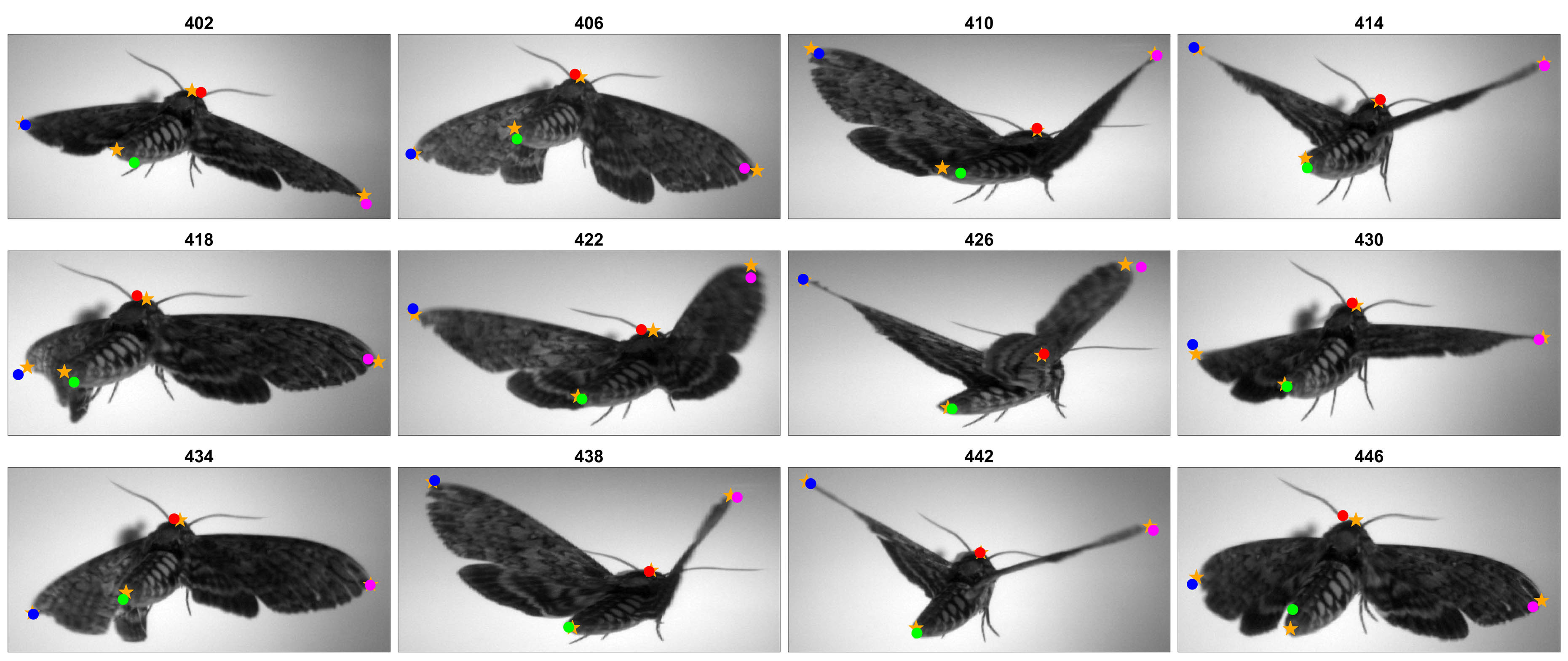}
	\caption{Training: Frames 1-400, Testing: Frames 401-800}
\end{subfigure}

\centering
\begin{subfigure}[!ht]{0.80\textwidth}
	\includegraphics[width=1.0\linewidth]{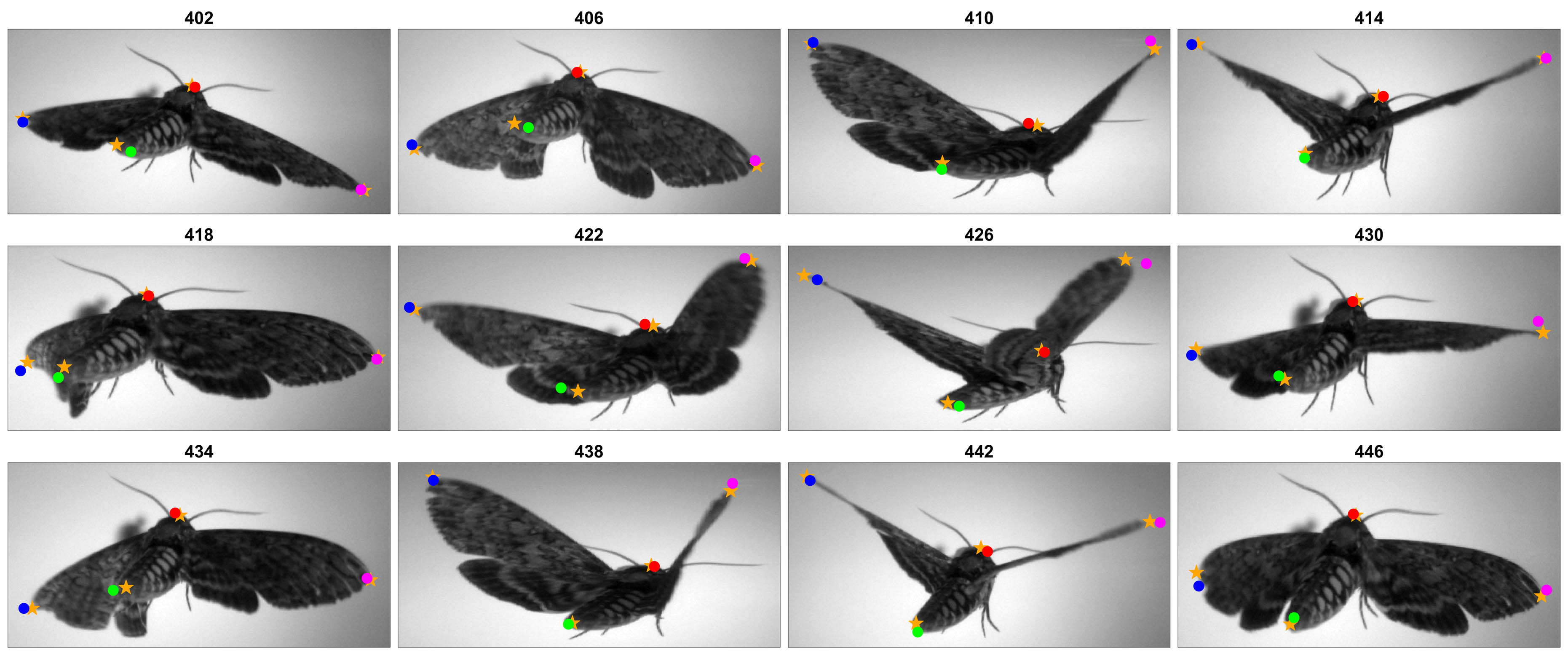}
	
	\caption{Training: Odd Numbered, Testing: Even Numbered}
\end{subfigure}

\caption[Quantitative and qualitative results demonstrating the influence of training and test split.]{\textbf{Quantitative and qualitative results demonstrating the performance obtained when using the default training split as compared to an interleaved one.} (a) Plot showing the mean absolute error associated with each landmark, computed for both splits, using a 400 frame test set. (b) Qualitative results are shown for 12 frames that are common to the test sets of both splits. Predicted landmark localizations for the default split are shown as colored circles and ground truth landmark localizations are shown as gold stars.  (c) Same type of plot as (b), but for the case of the interleaved split.}
\label{fig:comparison_split}
\end{figure*}

\begin{figure}[!ht]
\centering
	\includegraphics[width=0.9\linewidth]{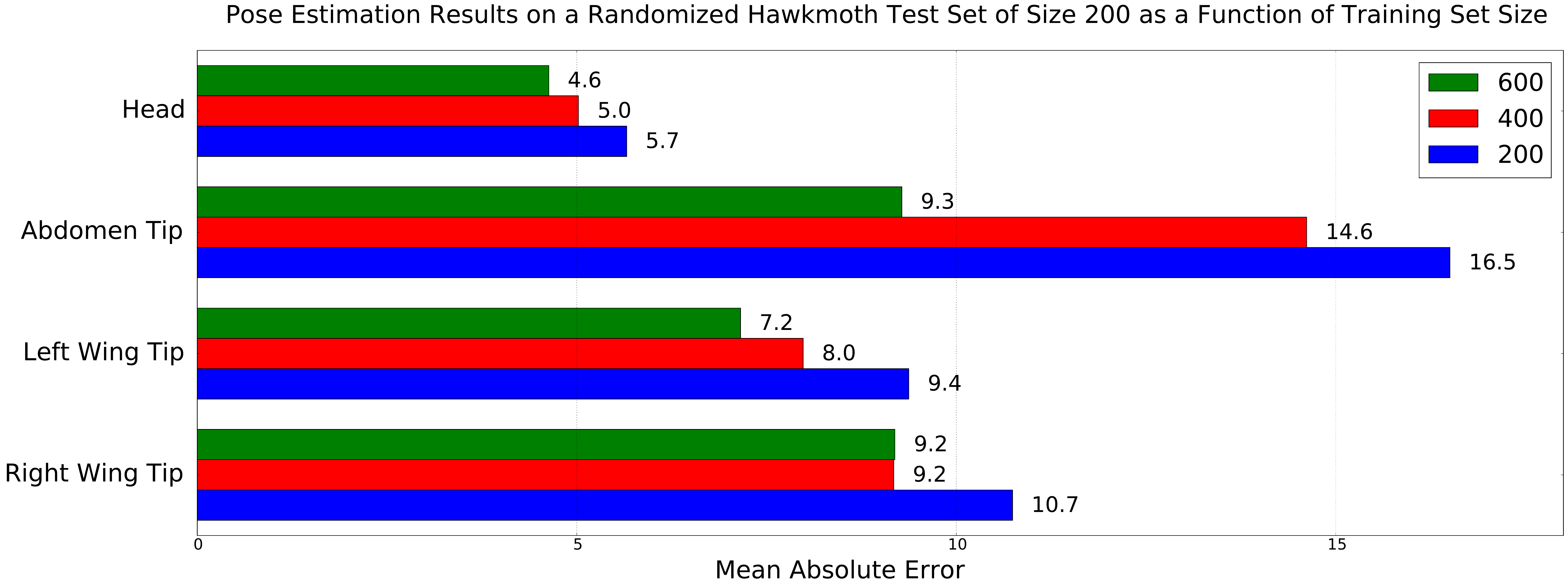}
\caption[Quantitative results demonstrating the influence of the original training set size.]{\textbf{Quantitative results demonstrating the influence of the training set size, prior to data augmentation, on test performance.} For each training set size, MAE is computed for each landmark type across a test set consisting of 200 randomly chosen images from an initial set of 800 frames. The training set sizes used in this experiment include 200, 400, and 600 images, which respectively represent 25\%, 50\%, and 75\% of the dataset.}
\label{fig:orig_training_size}
\end{figure}

\noindent \textbf{\normalsize Training and Test Split:} Using the default split resulted in a test loss of 142.7 and a training loss of 11.0. The alternative split, where the training and test sets interleave, resulted in a test loss of 33.6 with a training loss of 19.3. Figure \ref{fig:comparison_split} shows qualitative and quantitative results comparing these two splits. The quantitative results consist of the MAE associated with localizing each landmark. Figure \ref{fig:orig_training_size} shows quantitative results depicting how using training sets of size 200, 400, and 600 influence performance on a test set of size 200, measured with MAE. \\ 

\begin{figure*}[!ht]
\begin{subfigure}[!ht]{1.0\textwidth}
	\includegraphics[width=1.0\linewidth]{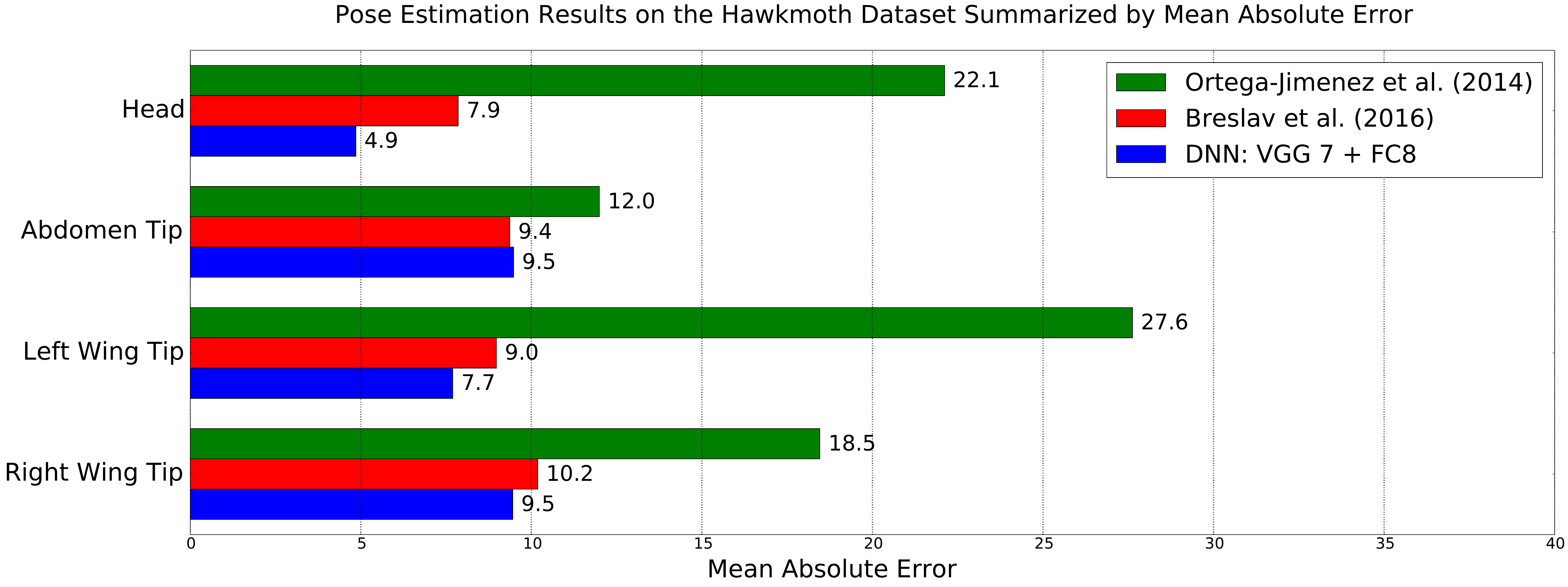}
	\caption{}
\end{subfigure}

\begin{subfigure}[!ht]{1.0\textwidth}
	\includegraphics[width=1.0\linewidth]{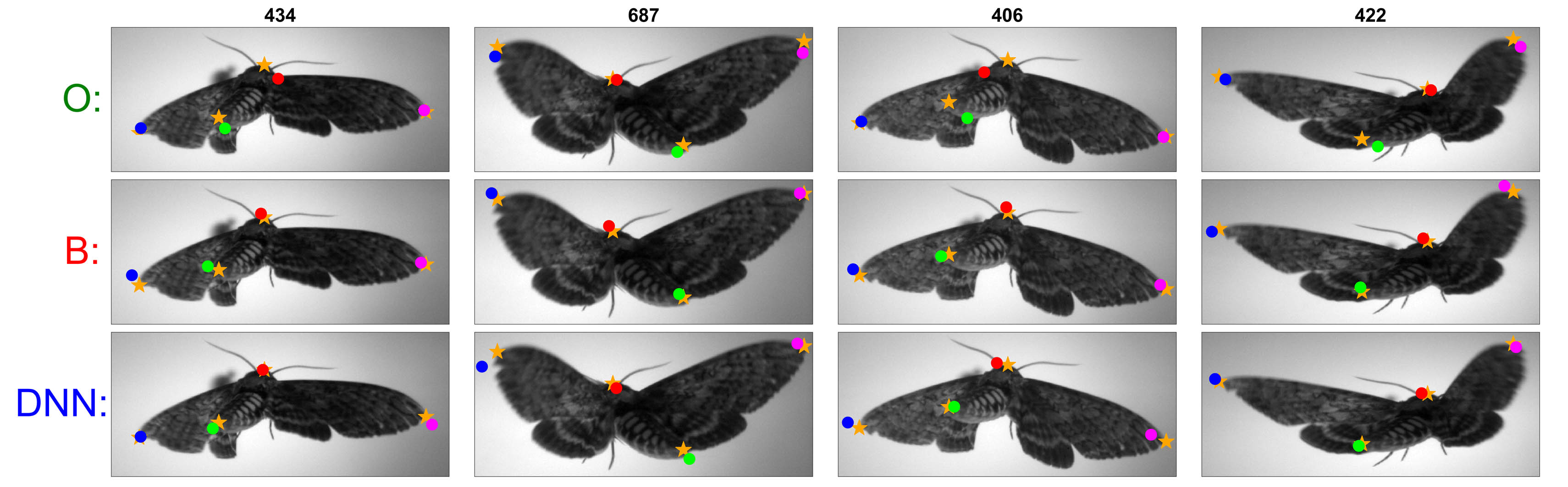}  \\
	\includegraphics[width=1.0\linewidth]{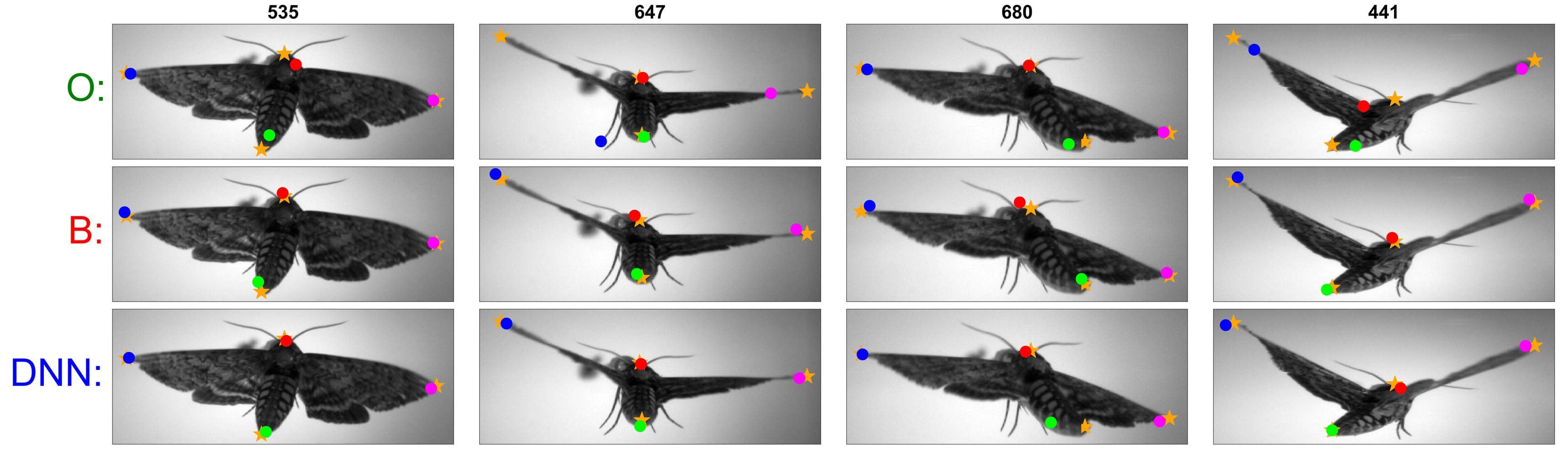}
	\caption{}
\end{subfigure}

\caption[Quantitative and qualitative comparison DNN with previous works on hawkmoth landmark localization.]{\textbf{Quantitative and qualitative comparison of our default DNN, denoted as VGG 7 + FC8, with the two best published approaches for hawkmoth landmark localization.} These approaches include the work of \cite{Ortega-Jimenez2014} denoted by ``O'' and \cite{Breslav2016} denoted by ``B''. (a) Plot showing the mean absolute error associated with each landmark, for each of the methods. (b) Plot showing the predicted landmark localizations (colored circles) along with ground truth landmark localizations (gold stars), for 8 different frames. Each column, labeled with a frame number, shows the performance of the three approaches on that frame.}
\label{fig:comparison}
\end{figure*}

\noindent \textbf{\normalsize Comparison to Previous Works:} Figure \ref{fig:comparison} shows how our default DNN, denoted VGG 7 + FC8, performs on automatic landmark localization on a hawkmoth dataset, relative to competing approaches whose performance was published in the work of \cite{Breslav2016}. Performance is measured for each landmark by computing MAE on the test set. The error for each landmark is defined as the Euclidean distance between the automatically generated landmark position and the corresponding ``ground truth'' (manually annotated) position. To be consistent with results reported in the work of \cite{Breslav2016}, we used the same training/test split and do not include frames where one or more landmarks were occluded in the quantitative results we report. Specifically, 421 images were used with 211 for training and 210 for test. \\ 

\begin{figure}[!ht]
\centering
	\includegraphics[width=1.0\linewidth]{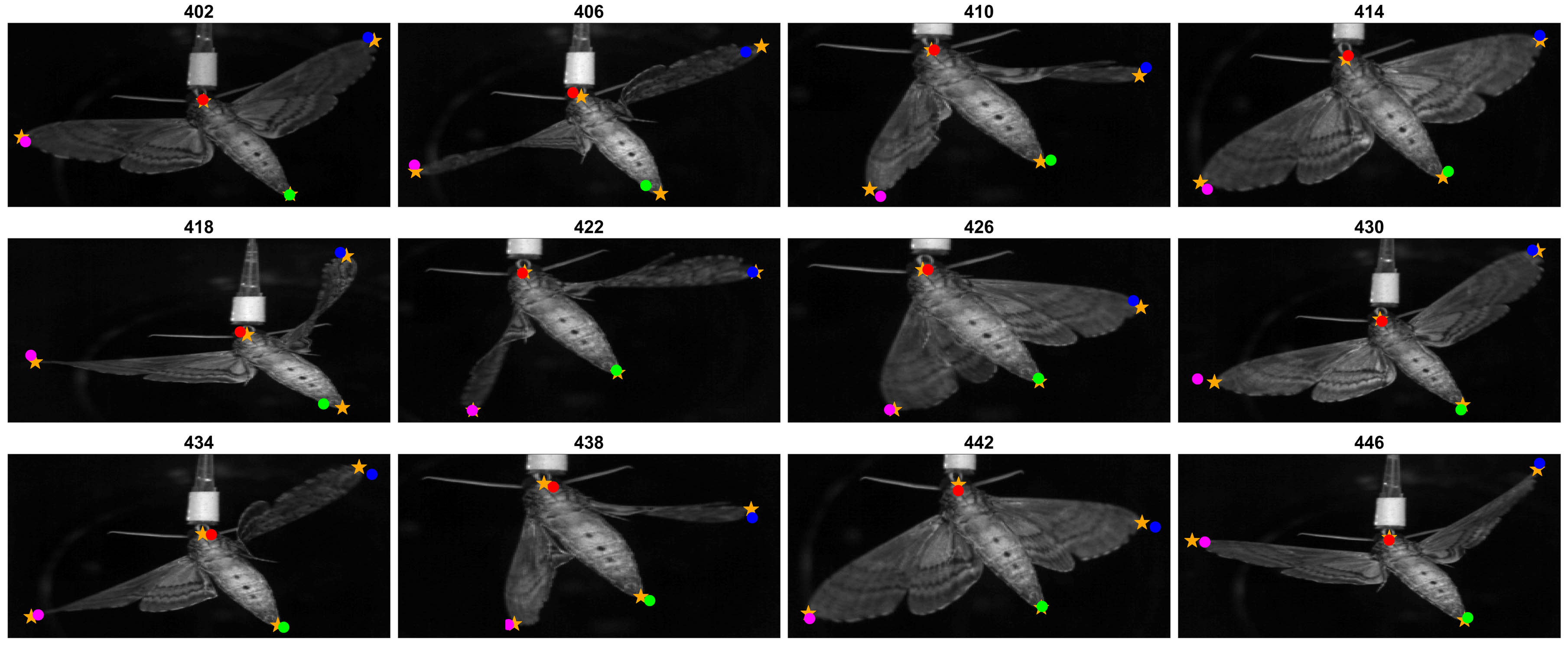}

\caption[Qualitative results of DNN on second camera view.]{\textbf{Qualitative results of DNN$_2$ on the test set from camera 2.} Results are shown for 12 different frames with the predicted landmark localizations (colored circles) and the ground truth landmark localizations (gold stars).}
\label{fig:cam2}
\end{figure}

\begin{figure*}[!ht]
\centering
	\includegraphics[width=0.8\linewidth]{3Dposes/402.png} \\
	\includegraphics[width=0.8\linewidth]{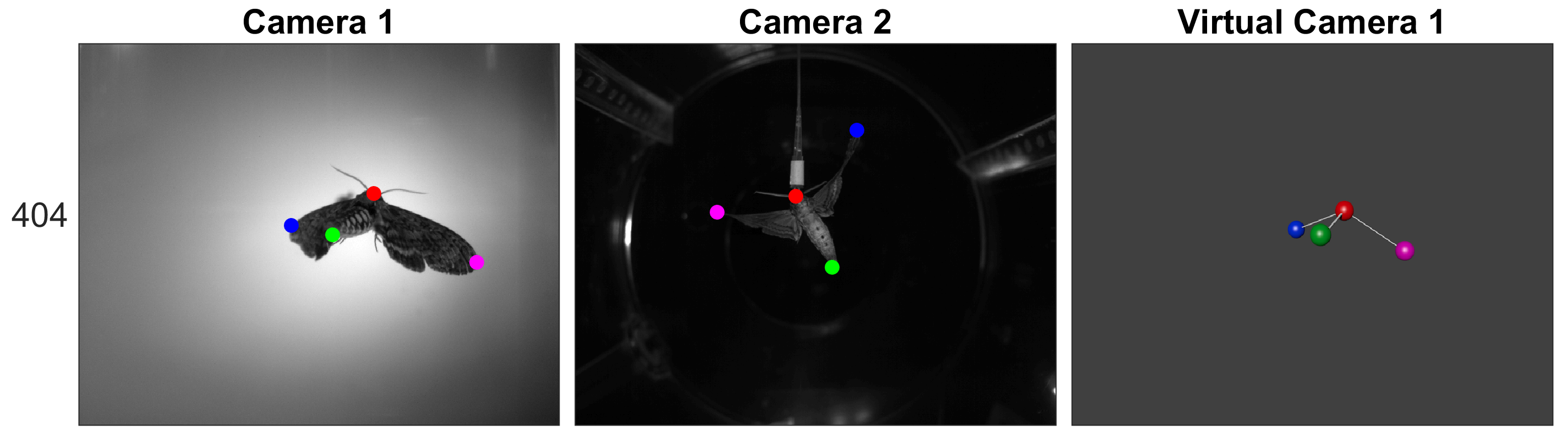} \\
	\includegraphics[width=0.8\linewidth]{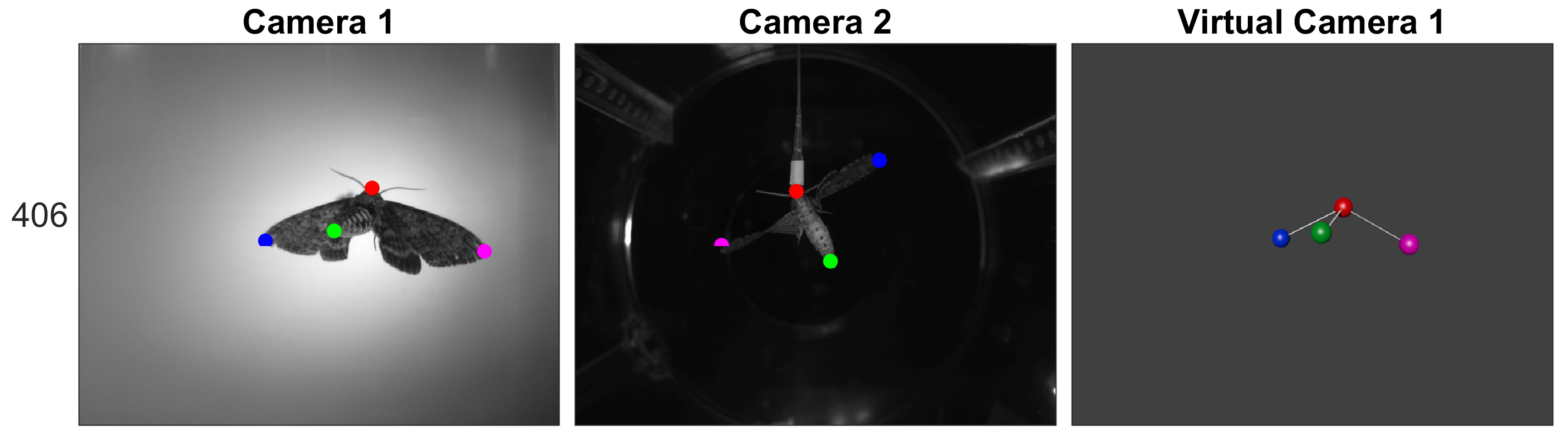} \\ 
	\includegraphics[width=0.8\linewidth]{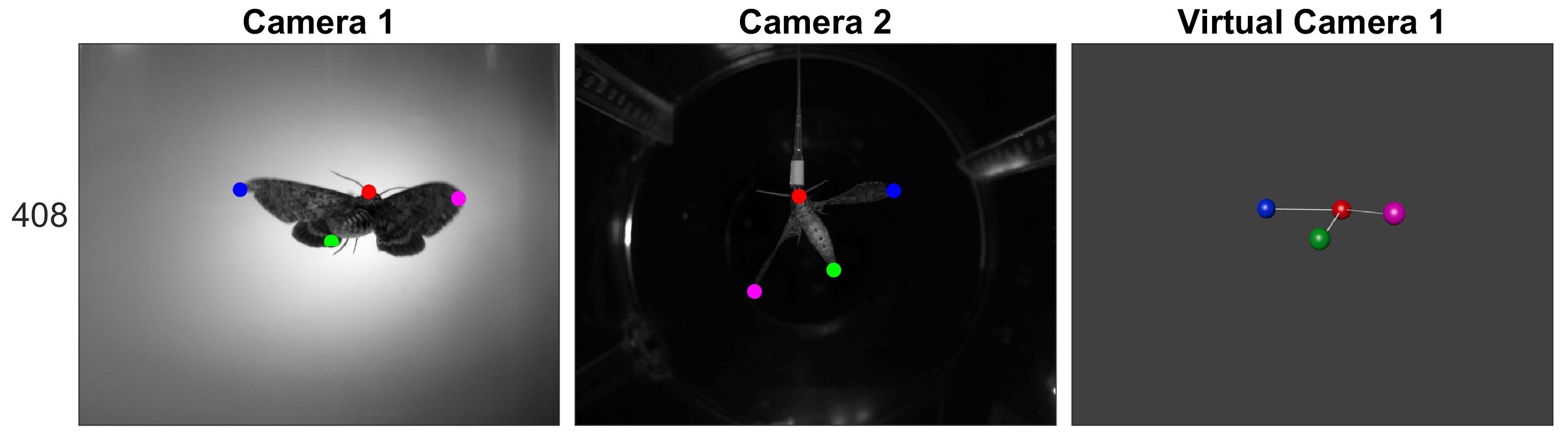} \\
	\includegraphics[width=0.8\linewidth]{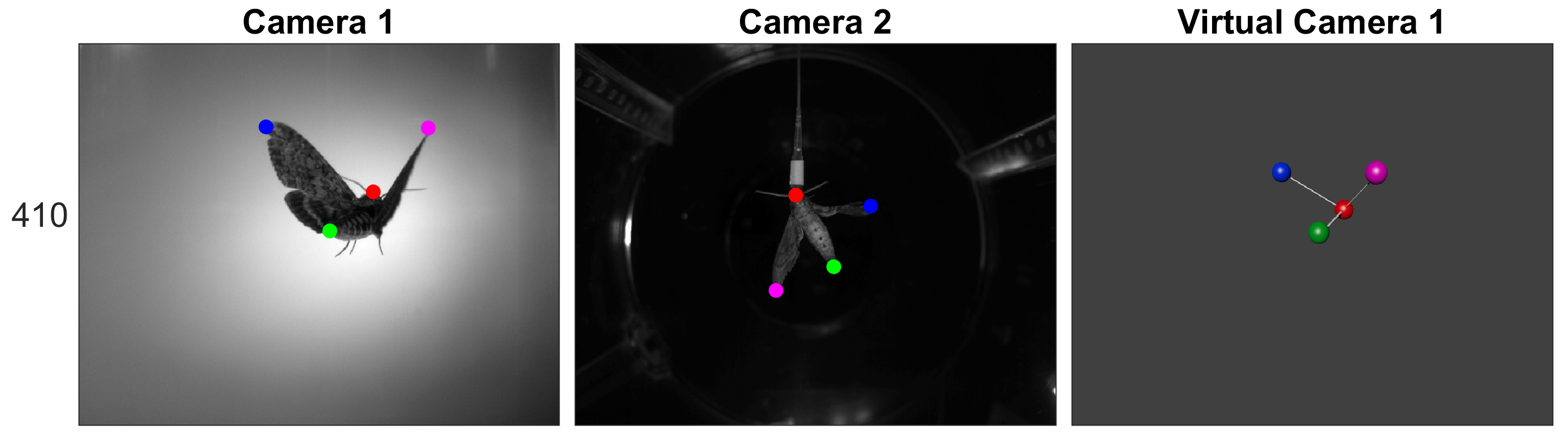}  \\
\caption[Qualitative 3D pose results.]{\textbf{Qualitative results of both DNN$_1$ and DNN$_2$ on landmark localization for frames simultaneously captured by both cameras 1 and 2.} The resulting landmark localizations are used to reconstruct the 3D positions of the landmarks. The right-most image in each subplot, labeled ``Virtual Camera 1'', illustrates the reconstructed 3D configuration (pose) of the hawkmoth. Results are shown for four frames with predicted landmark localizations denoted with colored circles and ground truth landmark localizations denoted with gold stars.}
\label{fig:multiview}
\end{figure*}

\noindent \textbf{\normalsize Multi-view:} DNN$_2$, our default DNN trained on video from camera 2 with the default train/test split, obtained a test loss of 191.8. Figure \ref{fig:cam2} shows the qualitative performance of DNN$_2$ on 12 different frames from the test set. Figure \ref{fig:multiview}, shows the qualitative performance of both DNN$_1$ and DNN$_2$ on landmark localization for frames simultaneously captured by both cameras 1 and 2. The resulting landmark localizations are used to reconstruct the 3D positions of the landmarks. The right-most image in each subplot, labeled ``Virtual Camera 1'', illustrates the reconstructed 3D configuration (pose) of the hawkmoth. 

To assess how accurate 3D reconstructed positions (based on our DNN predictions) are, we compare them with reconstructed 3D positions based on ground truth landmark annotations. Let $HA_i$ be the distance in 3D, at time $i$, between the predicted head and abdomen tip, and let $HA_i^*$ be the distance in 3D, at time $i$, between the ground truth head and abdomen tip. Similarly, let $LR_i$ be the distance in 3D, at time $i$, between the predicted left wing tip and the right wing tip, and let $LR_i^*$ be the distance in 3D, at time $i$, between the ground truth left wing tip and right wing tip. Then, the ratios $\frac{HA_i}{HA_i^*}$, and $\frac{LR_i}{LR_i^*}$, indicate at time $i$, how closely a 3D measurement of the distance between landmarks is to the ground truth distance, with a perfect match resulting in ratios of 1. Here we report the mean of these ratio taken over the test set. The mean ratio for the head to abdomen tip distance is 1.0096 and the mean ratio for the left wing tip to right wing tip distance is 0.9847.

\section{Discussion}
%

\noindent \textbf{\normalsize Number of Training Samples:} Our experiments show that increasing the number of training samples from $10^{3}$ to $10^{4}$ has a significant impact on reducing the test loss from approximately 400 to approximately 150. Beyond $10^{4}$ samples the test loss stays around 150, indicating that the benefit of additional training samples diminishes. Recall that these training samples are a result of performing data augmentation with translations on an initial set of 400 samples. This suggests that creating a training set that is 25 times larger than the original dataset, using data augmentation with translation alone, is sufficient. It is worth emphasizing that these results do not tell us how changes in the original number of training samples would impact the results. The trends observed in this experiment agree with the idea that when the training set is relatively small ($10^{3}$), it is easier for a DNN to learn or potentially memorize the data (also referred to as ``overfitting''), leading to a small training loss. This overfitting comes at the expense of the ability of the network to generalize which is reflected in larger test losses. The saturation in test loss as the number of training samples goes beyond $10^{4}$ also appears to be consistent with our hypothesis that the benefit of data augmentation is limited and cannot replace more original training samples. \\

\noindent \textbf{\normalsize DNN Architecture:} Our experiments show that the feature extraction portion (VGG $X$) of the DNN makes a difference in test performance, but not as much as hypothesized. Our initial hypothesis was that the best performing architecture would be the one which uses the lowest level features (VGG 2 + FC8), and the worst performing architecture would be the one which uses the highest level features (VGG 13 + FC8), and that the difference between the two would be large. The rationale for this hypothesis is that the VGG layers are pretrained on ImageNet which contains many different scenes and objects that at a high level have little in common with the hawkmoth dataset, and so only the lowest level features would be useful to our application. Our hypothesis was incorrect as the best performing architecture was VGG 7 + FC8, which can be interpreted as an architecture which uses features that are not as low level as VGG 2 + FC8, nor as high level as VGG 13 + FC8. Consistent with our hypothesis the worst performing architecture was VGG 13 + FC8. Surprisingly, when comparing across learning rates the difference in test loss between the best performing architecture and the worst is only approximately 50. 

One potential reason why the architecture may not make as large of a difference as hypothesized is that the features computed by the VGG $X$ portion of the network are hierarchical in nature. This means that features output in one layer depend on the features output in earlier layers. If the low level features alone are sufficient for learning a linear regression model that accurately localizes landmarks then higher level features which depend on these low level features may preserve enough low level information to also be of use. In this case, the FC8 fully connected layer is able to learn a linear regression model that works across different feature spaces.

We also performed experiments to determine whether adding an additional fully connected layer to VGG 7 + FC8 would improve test performance by learning a non linear regression model. The result of these experiments was that the training loss converged to values that were too large to be of any use. Our results here also support the idea that training success is strongly dependent on the learning rate. For each network studied we found training loss diverged once the learning rate was set too large.  \\



\noindent \textbf{\normalsize Network Initialization:} Our experiments show that when training a network from scratch ``xavier'' initialization greatly outperforms Gaussian (0 mean, 0.01 standard deviation) initialization. Still, training a network from scratch results in inferior performance compared to using pretrained VGG $X$ layers. These results support the hypothesis that a small training set, even augmented, is insufficient for learning quality feature maps. There maybe no choice for small datasets but to rely on pretrained networks which allow one to leverage feature maps learned on large datasets \citep{Razavian2014}. \\

\noindent \textbf{\normalsize Number of Iterations:} Our experiments show that performing 4000 iterations of training was enough to achieve a loss close to the local minimum, and that additional training did not significantly lower the loss. Relative to the default batch size of 32, 4000 iterations results in the network seeing 128,000 training samples during training. At 6250 iterations the network would perform 1 ``epoch'' meaning it would see the whole training set of 200,000 samples once. \\

\noindent \textbf{\normalsize Data Augmentation:} Our experiments show that performing data augmentation with translation only outperforms combinations of translation with rotation and translation with scale. Our initial hypothesis was that by adding rotated and scaled versions of the hawkmoth to the training set, the network would be forced to learn a more general representation of hawkmoth landmarks, which would result in better generalization and performance on test images. In practice, our test set did not have significant in-plane rotations or scale changes compared to the original (un-augmented) training set, so we believe that augmentation with rotation and scale acted as a distraction to the network which ultimately hurt performance on the test set. We also want to emphasize that data augmentation is absolutely crucial to preventing our architectures from overfitting to the training set, as shown by the enormous loss that results from no data augmentation being performed. \\

\noindent \textbf{\normalsize Batch Size:} Our experiments show that increasing the batch size results in a lower test loss, but has diminishing returns beyond a batch size of around 32. These results agree with the notion that a larger batch size allows for a more accurate approximation of the gradient of the loss function with respect to the weights and biases of the network, which allows SGD to descend in a direction closer to the optimal direction. However, the time it takes to train a network is linear in the batch size, so a doubling in the batch size will result in a doubling of the training time. To train a DNN with a batch size of 128 took more than 11 hours. We also performed an experiment with a batch size of 8 which resulted in a diverging loss during training. \\

\noindent \textbf{\normalsize Training and Test Split:} Our experiments show that an alternative split, where the training and test sets interleave, results in a lower test loss. This supports the hypothesis that the more similar the training and testing distributions are, the closer the training loss will be to the test loss. In our application of landmark localization, given a fixed size training set, it is clearly favorable to choose the images in the training set so they overlap as much as possible with the testing data.

Another experiment we performed addresses the question of how increasing the size of the original training set, prior to any data augmentation, influences performance on a test set containing 200 randomly chosen images. Our hypothesis is that the more original training data available the better the test performance will be. The results mostly confirm this where the test loss decreases as the original training set size increases. An interesting question that arises is how much of a difference does it make when using a training set of 200 images or 25\% of the original dataset, as compared to 600 or 75\% of the original dataset? Our experiments reveal that for some landmarks the difference can be as little as approximately 1 pixel in terms of MAE, or as large as approximately 7 pixels in terms of MAE. The abdomen tip in our experiment is the landmark whose localization performance benefits the most from additional training data. This can be explained by the fact that the pose of the abdomen tip varies significantly across the video making it more difficult to learn a model as compared to other landmarks. Our results suggest that annotating as little as 25\% of a dataset can be sufficient to localize some landmarks reasonably well, but other landmarks may have unacceptably high error. Increases in the training set size do help but must be weighed against the cost of producing more annotated training data.\\

\noindent \textbf{\normalsize Comparison with Previous Works:} The focus of our experiments has largely been on studying the impact of parameter values on testing loss. Now we aim to put the results obtained with our default DNN (VGG 7 + FC8) into context. One way we contextualize our results is by comparing them to already published results on hawkmoth landmark localization \citep{Breslav2016}. The results presented in Figure \ref{fig:comparison}a show the MAE that our default DNN achieves per landmark relative to two other approaches. Our default DNN outperformed the next best approach on localizing the head, left wing tip, and right wing tip. However, for the abdomen tip our DNN is outperformed by the approach of \cite{Breslav2016}. As an overall measure of performance for an approach we compute the sum of MAEs across landmark types. This resulted in the method by \cite{Ortega-Jimenez2014} obtaining a total MAE of 80.2, \cite{Breslav2016} obtaining a total MAE of 36.5, and our default DNN obtaining a total MAE of 31.6. The conclusion here is that our proposed default DNN is the best performing among published works for this application.

The landmark that is easiest for our DNN to localize is the head, then followed by the left wing tip, abdomen tip, and the right wing tip. It is not surprising that the head is easiest to localize since its position and orientation relative to the camera are the least variable due to the hawkmoth being engaged in feeding during the video capture. Specifically the hawkmoth is feeding on artificial nectar contained in a plastic pipette tip that is fixed in position. The abdomen tip is expected to be difficult to localize partly because it does not have a distinct appearance compared to say the abdomen or other more textured regions. It is not clear why the right wing tip is more difficult to localize than the left wing tip, but overall the wing tips are also expected to be relatively difficult to localize due to the large changes in position they undergo during flight. It is also worth noting that the quantitative results discussed here do not include occlusion cases. However, we have labeled occluded landmarks in our training set and we have observed that our DNN is able to learn to predict the position of occluded landmarks. \\

\noindent \textbf{\normalsize Multi-view:} Our multi-view experiments are included to emphasize one of the original motivations of our work: the quantitative study of 3D flight kinematics of a flying animal. In Figure \ref{fig:multiview} we demonstrate the result of two DNNs (DNN$_1$ and DNN$_2$) automatically annotating landmarks in their respective camera views, and how the 3D positions of the landmarks can subsequently be reconstructed. The ratios we have computed in the results section are very close to 1 and suggest that using the output of our DNNs to measure quantities in 3D will result in measurements that are very close to ground truth. Looking at Figure \ref{fig:multiview} we note that the virtual camera subplot shows the 3D flight of the hawkmoth. This use case is very exciting as our DNNs can enable scientists to obtain more accurate 3D positions of flying animal landmarks in less time, ultimately leading to analyses that better reflect the true kinematics of the animal. \\

\noindent \textbf{\normalsize Recommendations:} Given our experimental results we make the following recommendations to our readers who would like to train DNNs for their own data and application. 

\begin{enumerate}
\item Use as large of an initial training set as possible and definitely perform data augmentation with translation alone to obtain a training set that is about 25 times larger than the original set. Consider other data augmentation types based on the invariances you want the network to learn.

\item As a starting point, use the architecture VGG 7 (pretrained) with a single fully connected layer attached. The number of neurons in the fully connected layer should be set to twice the number of landmarks that you wish to localize (a single neuron per dimension). 

\item Choose the learning rate to be the largest value that still allows the training loss to converge.

\item Choose the number of training iterations by taking some of your training set and using it as a validation set (analogous to how our test set has been used here). Set the number of iterations to the value where validation set loss stops decreasing significantly. For a quick starting point try 10,000 iterations. 

\item A batch size of 32 is a good starting point as it balances performance, accuracy of the gradient computation, and total training time. 

\item Choose the training set so that it is as similar to the images that need to be annotated as possible. 

\end{enumerate}

\section{Conclusion}
Our work presents a comprehensive study on how numerous parameters influence the training and testing performance of DNNs for the application of automatic annotation of landmarks. Our experiments show that DNNs can be used for landmark localization and that our DNN ``VGG 7 + FC8'' outperforms leading approaches on the problem of landmark localization in hawkmoth videos. We demonstrate how DNNs applied to multiple camera views can enable estimation of the 3D pose of flying animals. Additionally, we facilitate the use of DNNs by researchers from different fields by providing a self contained explanation of what DNNs are, how they work, and how they can be applied to other datasets using the freely available library Caffe together with our freely available example code\footnote{\url{http://www.cs.bu.edu/~betke/research/ALADNN/}}. Furthermore, to help support ongoing research on landmark localization we are making freely available a new dataset, called HRMF2\footnote{\url{http://www.cs.bu.edu/~betke/research/HRMF2/}}, consisting of high resolution multi-view data of hawkmoths along with landmark annotations and segmentations. To the best of our knowledge this would be the first published, freely available dataset of its kind.

\section{Acknowledgments}
We wish to thank University of North Carolina undergraduate Hanna Gardner for helping to annotate moth datasets. We also thank the Image and Video Computing group at Boston University for their valuable feedback.

%
%

%

\section{Funding}
This work was partially funded by the Office of Naval Research [N000141010952 to M.B. and T.H.], the National Science Foundation [0910908 and 0855065  to M. B. and S. S. and 1253276 to T. H.] and the  Air Force Office of Scientific Research [FA9550-07-1-0540 to M. B.].

\section{Data availability}
Hawkmoth data is available at \url{http://www.cs.bu.edu/~betke/research/HRMF2/}. Example code facilitating the training and testing of DNNs is available at \url{http://www.cs.bu.edu/~betke/research/ALADNN/}. 

\newpage
\section{Appendix A}
\label{sec:appendixa}

Here we provide a more in depth explanation of the different types of layers used in the VGG 16 network, which are also common to many other CNNs. \\

\noindent \textbf{\small Convolutional:} A convolutional layer is a volume of neurons with the special property that individual neurons only connects to a subset of neurons in the previous layer, and neurons at the same depth in the volume share the same weights and biases. As an example, the first convolutional layer of VGG 16 has a height of 224 neurons, a width of 224 neurons, and a depth of 64. Each neuron in the first convolutional layer is connected with a 3 x 3 pixel region in each of the 3 channels (RGB) of the input image. As a result a single neuron in the first convolutional layer must learn 3 x 3 x 3 weights and 1 bias. Since all neurons at the same depth (depth slice) share the same weights and bias, then each depth slice requires learning only 27 weights and 1 bias, resulting in a total of 1728 (27 x 64) weights and 64 (1 x 64) biases for the whole layer. Conceptually, each depth slice is learning a filter that is sensitive to some pattern. Collectively this means the first convolutional layer learns 64 filters. The second convolutional layer in VGG 16 also has a height of 224, width of 224, and depth of 64, but now each neuron connects to a 3 x 3 region of each of the 64 depth slices in the previous layer, resulting in 3 x 3 x 64 weights and 1 bias per neuron. In Figure \ref{fig:vgg16}, the convolutional layers of VGG 16 are depicted as black rectangular volumes with a label `Conv,' short for convolution, and a number indicating the depth of the volume. Notice that layers towards the end of the network (output side) have larger depths but smaller heights and widths. \\

\noindent \textbf{\small Fully Connected:} A fully connected layer is a layer where each neuron is connected with all outputs from the previous layer. In VGG 16 there are 3 fully connected layers, shown in blue in Figure \ref{fig:vgg16}. The first fully connected layer has 4,096 neurons each of which connect to all outputs from the previous layer. Since neurons in the fully connected layers connect to all outputs of the previous layer, instead of just a subset, fully connected layers account for a large percentage of the weights and biases present in a typical CNN. \\

\noindent \textbf{\small Non Linearity:} As previously mentioned, a neuron applies a non linear function to a weighted combination of inputs. The choice of function to apply is a design decision. Since each neuron applies a non linearity, it is natural to consider it as a parameter of the convolutional layers and also the fully connected layers. This is the case in Figure \ref{fig:vgg16} where the non linearity is not explicitly shown. However, to make the choice of non linearity more explicit, it can be thought of as its own layer. From this point of view each convolutional and fully connected layer are immediately followed by a non linearity layer. The output dimension of the non linearity layer will be the same as its input dimension. One popular choice for non linearity is the rectified linear unit (ReLU), defined as: $f(x) = max(0,x)$. The ReLU is used in VGG 16 and was shown to be advantageous for training deep networks \citep{Glorot2011}.\\  

\noindent \textbf{\small Pooling:} Pooling layers apply a pooling function to reduce the dimensionality of the input. A common pooling function is max pooling where the output of a node is the maximum over a region from the input. In VGG 16, max pooling is performed over a 2 x 2 window, with a ``stride'' of 2 which means that an input volume of size 224 x 224 x 64 will be transformed into an output volume of size 112 x 112 x 64. In Figure \ref{fig:vgg16} all max pooling layers are denoted in red. \\

\noindent It is worth noting that the VGG 16 network shown in Figure \ref{fig:vgg16} also has a final layer in orange named `Soft Max.' This soft max layer facilitates the computation of a loss function by remapping the outputs of the last fully convolutional layer to class probabilities between 0 and 1. Figure \ref{fig:vgg16} illustrates not only the architecture of the VGG 16 network but also the action of the network on an input image of a person. The network output is a list of probabilities, one for each class, and as desired the class ``person'' has the highest probability.

\section{Appendix B}
Here we describe how to accomplish two tasks using the deep learning library Caffe \citep{Jia2014}. The first task is to train or finetune a DNN using some training data. The second task is to run a trained DNN on some testing data to get output. To facilitate these two tasks we provide example code that can freely be used\footnote{\url{http://www.cs.bu.edu/~betke/research/ALADNN/}}.

\subsection{Training Set Preparation}
Before training or finetuning a DNN one needs to prepare a training set and ensure it is formatted properly for use with Caffe. A training set should consist of training samples and their labels. Since the focus of this work is on image annotation, we will think of training samples as individual images. The label associated with an image depends on the task that we are training the DNN to perform. In the case of classification where there are $N$ classes, the label should be a single integer in the range of $0$ and $N-1$ inclusive. For a multi-label classification problem where $K$ labels are predicted per training sample, the overall label will be a vector of dimensionality $K$, with each dimension containing an integer indicating the true class for that label. In the case of regression problems the label is continuous valued and can either be a single value, or a vector of values for the case of multiple regression. Example: The training set used in our experiments consists of images of a Moth where the label for each image is an 8 dimensional continuous valued vector containing the x and y coordinates of four landmarks. 

\subsubsection{Preprocessing and Data Format}
When training or finetuning a DNN, the training images need to conform to the architecture of the DNN. The first layer of the DNN will determine what the input to the network should be. To obtain training images that are suitable for input it is common to perform one or more preprocessing steps. In the case of the VGG 16 network, all input images needs to have 3 channels with a width and height of 224 pixels. Images that have a different width and height should be resized to $224 \times 224$. In the case that the images is a single channel or grayscale image, one can create a 3 channel images by replicating the original image in each channel. If finetuning a pre-trained network, additional preprocessing steps may need to be performed. For example, if the original network was trained on 8 bit images, one should ensure their images are also 8 bit. Additionally, it is common to normalize the input image values by subtracting a mean value. To finetune VGG 16, the authors provide per channel (blue, green, red) constants which are to be subtracted from each corresponding channel in the input. \\

\noindent For Caffe to be able to use one's training data, the data should be stored in two 4D arrays, one containing the training images, and the other containing the labels. These two 4D arrays are then written to disk as an HDF5 file. The 4D array storing images should have dimensions $N \times C \times H \times W$, where $N$ is the number of training images, $C$ is the number of channels per image, $H$ is the height of the image, and $W$ is the width of the image. The same format is used for the labels, but $C$ and $H$ will be 1, with the last dimension $W$ containing the label. Example: In our work the data 4D array has dimensions $1000 \times 3 \times 224 \times 224$, and the label has dimensions $1000 \times 1 \times 1 \times 8$. Note that here $N$ is 1000, but the total number of training samples we use is much larger. The reason one should not put all training samples into a single 4D array is because that would create a single file that is enormous. Instead the training set is broken into smaller batches, and each batch is represented by a pair of 4D arrays that are written to a single HDF5 file. Thus in our example, our 100,000 training samples are broken into 100 batches each containing a pair of 4D arrays which store $N=1000$ training images and their labels. Since each batch is written to a separate HDF5 file, we obtain 100 HDF5 files.  \\

\noindent The data format for the 4D arrays previously described is consistent with how Caffe thinks of data. However, when writing to HDF5 files the 4D arrays need to be modified so that the order of their dimensions is reversed, e.g $W \times H \times C \times N$. After the 4D data and label arrays are in this format they can be written to disk using an appropriate library that supports writing HDF5 files. In our work we use the built in Matlab functions h5create and h5write for this,  however libraries for Python and other languages should also be readily available. HDF5 files can have different fields in the same file, so in our case a single HDF5 file will have a /data field storing the training image 4D array, and the /label field storing the label 4D array. For specifics refer to our example code moth\_dataset\_augmentation.m in the supplemental materials. Note that the reversal of dimensions of the 4D arrays is an extra step that strictly speaking is not required but it makes it easier to differentiate the format used by Caffe with the one used by HDF5. \\

\noindent Lastly, it is worth remembering that data augmentation can be beneficial for training a DNN. Any data augmentation that a user wishes to perform on an original data set should be done \textit{prior} to the preprocessing and data format steps above. Note that with relatively large datasets the storage for HDF5 files can be large. For example 100 HDF5 files, each containing 1000, $224 \times 224  \times 3$ images takes up about 100 GB.  

\subsection{Training a DNN with Caffe}
Training a DNN with Caffe requires the creation of 3 text files which collectively specify the input data, the network architecture, and the training and optional testing parameters. \\

\noindent The main text file called train\_val.prototxt contains a specification of the DNN architecture, layer by layer, starting from the input data and ending at the loss function. Each layer is denoted in the text file with the word layer followed by an open and close parentheses. Within the parentheses is a specification of the type of layer and any pertinent parameter values. The first layer in this file will be a data layer, and in the case of HDF5 files, it will have a type of ``HDF5Data''. In this case the layer will need to reference a text file that contains the file paths of all the HDF5 files used for training. Layers with learnable parameters, like the convolutional and fully connected layers, will have text specifying the learning rate and initial values for the weights and biases in that layer. The second text file called deploy.prototxt is the same as train\_val.prototxt but with the data and loss layers removed. See our example code in the supplemental material. \\

\noindent So far we have made the assumption that ``training'' means training all of the layers of a DNN from scratch, where all weights and biases are initialized by the user. There are two variations to this kind of training which are also useful in practice, and require some modifications to the train\_val and deploy text files. The first variation is when a user only wants to train \textit{some} of the layers in the network. This can be accomplished by setting the learning rate of all other layers to zero. A second variation occurs when a user wants to initialize the weights and biases of some layers of their network to the weights and biases of an already trained network. This commonly occurs when someone wants to take a piece of a network that has already been trained and add their own layers on top of it. To initialize the weights and biases of some layers in the network to values learned from another network, one should not include any weight or bias initialization in the train\_val or deploy text files. Additionally, the name of the layers, which are to be initialized in this way, should match those used in the already trained network. Lastly, the weights and biases of the already trained network will be stored in a .caffemodel file which will be specified as an additional argument during training. See supplemental materials for an example.\\   

\noindent The third text file called solver.prototxt, specifies training parameters, and optional parameters if one wishes to evaluate the currently trained model on a test set, as the training process goes on. The solver text file will reference the train\_val.prototxt file, so that the solver knows what data and architecture to use for training. One of the most important parameters in the solver file is the base learning rate. The base learning rate is multiplied by the learning rate values (multipliers) specified in the train\_val.prototxt to obtain the final learning rate used for any layer. For example if the base learning rate in the solver is $10^{-5}$ and the learning rate specified in a certain fully connected layer is $100$ then the learning rate used for that fully connected layer will be $100 \times 10^{-5} = 10^{-3}$. Other parameters in the solver file determine how the base learning rate is changed (its ``schedule'') as training proceeds. The total number of training iterations to perform is specified by the parameter max\_iter. As training occurs it is wise to have Caffe save the weights and biases of the network to disk in the form of a .caffemodel. The parameter snapshot specifies that every snapshot iterations a .caffemodel will be written to disk. If all goes well the very last .caffemodel written is the one that contains the weights and biases learned after max\_iter iterations of training. See supplemental materials for an example. \\

\noindent After these 3 text files are created and specified, Caffe can be run from the command line. The following is an example call: {\code caffe train -gpu 0 -solver solver.prototxt}. If you want to initialize some of the weights and biases in the network to those that were learned for another network you would use the following command line command: {\code caffe train -gpu 0 -solver solver.prototxt -weights name.caffemodel}. Additional details on layer types, solver parameters, training and testing usage from the command line, are available on the Caffe website\footnote{http://caffe.berkeleyvision.org/tutorial/}.


\subsection{Testing a DNN with Caffe}

The Caffe library provides interfaces for Matlab (MatCaffe) and Python (PyCaffe), and either can be used to run a trained DNN on test data. In our work we use the MatCaffe interface and create a simple Matlab script to run our trained DNNs on test data. The script creates a caffe network object which is initialized by specifying a deploy.prototxt file and the .caffemodel containing the weights and biases learned during training. This object, which we call net, has the forward function called which takes in a cell array of size 1, containing a 3D array of dimension $W \times H \times C$, which contains the test data. The output of the function is a cell array of size 1 containing the output of the network, which is just the input forward propagated through the network. The output can be used for evaluating performance and visualizing the result of the network. Note that any preprocessing steps performed on the training data, should also be performed on the test data. Typically, testing is fast since a single input only requires a single forward pass over the network to obtain an output. See supplemental materials for an example of testing.

\pagebreak

\bibliography{mybibliography} 
\bibliographystyle{jeb}

\end{document}